\documentclass[11pt,a4paper]{article}
\usepackage[hyperref]{emnlp2020}
\usepackage{times}
\usepackage{latexsym}
\usepackage{amsmath}
\usepackage{amssymb}
\usepackage{pifont}
\usepackage{multirow} 
\usepackage{paralist}
\usepackage{enumitem}

\usepackage{microtype}
\aclfinalcopy 

\newcommand{\anonymous}[1]{#1}

\usepackage{cprotect}  

\usepackage{graphicx}
\usepackage{comment}

\newcommand{\cmark}{\ding{51}}%
\newcommand{\xmark}{\ding{55}}%

\title{STAR: A Schema-Guided Dialog Dataset for Transfer Learning}
\newcommand{\DATASETNAME}{STAR}
\newcommand{\DATASETURL}{\url{https://github.com/RasaHQ/STAR}}

\author{Johannes E.\ M.\ Mosig\thanks{  ~~Equal contribution} \\
  Rasa \\
  \texttt{j.mosig@rasa.com} \\\And
  Shikib Mehri\footnotemark[1] \\
  Language Technologies Institute \\
  Carnegie Mellon University \\
  \texttt{amehri@cs.cmu.edu} \\\And
  Thomas Kober \\
  Rasa \\
  \texttt{t.kober@rasa.com} \\}

\date{}

\begin{document}
\maketitle
\begin{abstract}
We present \DATASETNAME, a schema-guided task-oriented dialog dataset consisting of 127,833 utterances and knowledge base queries across 5,820 task-oriented dialogs in 13 domains that is especially designed to facilitate task and domain transfer learning in task-oriented dialog.
Furthermore, we propose a scalable crowd-sourcing paradigm to collect arbitrarily large datasets of the same quality as \DATASETNAME.
Moreover, we introduce novel schema-guided dialog models that use an explicit description of the task(s) to generalize from known to unknown tasks. 
We demonstrate the effectiveness of these models, particularly for zero-shot generalization across tasks and domains.
\end{abstract}

\section{Introduction}
\label{sec:introduction}

A long-standing challenge in computer science is to develop algorithms that can interact with human users via dialog in natural language~\citep{turing1950computing,DSTC8}. 
Of particular interest is task-oriented dialog, wherein a user interacts with a system to achieve some goal (e.g.\ making a doctor appointment). 
The system should understand the user's requests and assist them by taking the appropriate actions (e.g.\ searching a database or replying to the user).
In recent years, supervised learning approaches to this problem have become particularly popular~\cite{gao2018neural}, because they can potentially learn complex patterns without relying on hand-crafted rules.
While such data-driven methods already demonstrate impressive performance in open-domain dialog \citep{zhang2019dialogpt,adiwardana2020towards,roller2020recipes}, task-oriented dialog models face the additional difficulty of transferring skills to tasks and domains that were not present in the training data. 
To address this issue, we present the \textbf{S}chema-guided Dialog Dataset for \textbf{T}ransfer Le\textbf{ar}ning (\textbf{\DATASETNAME}) dataset, a collection of realistic, task-oriented dialogs, that is especially designed to test and facilitate the transfer of learned patterns between tasks.

\begin{figure}[htb]
    \centering
    \includegraphics[width=0.95\columnwidth]{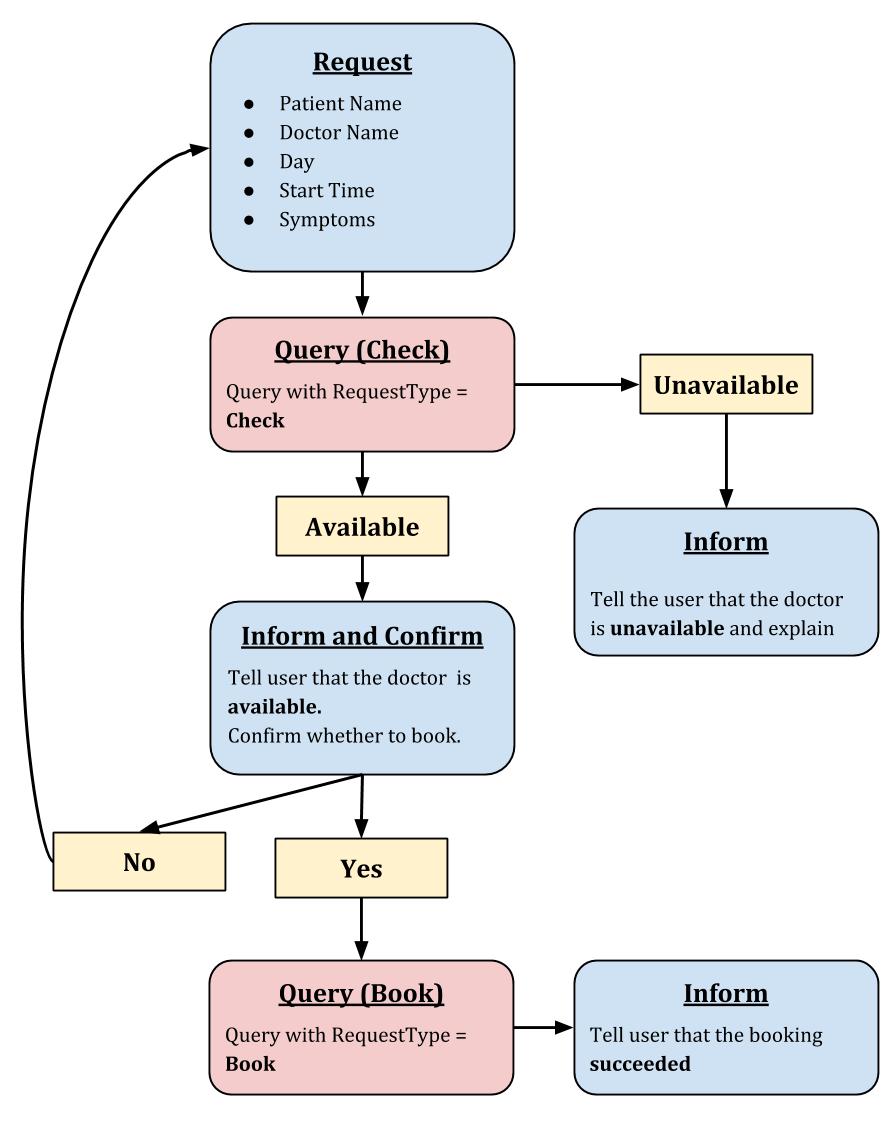}
    \cprotect\caption{%
        Flow chart representation of the schema-graph for the \verb+book_doctor_appointment+ task.
        The corresponding schema file is shown in Appendix~\ref{apx:format}.
    }
    \label{fig:doc_schema}
\end{figure}

Unlike open-domain dialogs, task-oriented dialogs are accompanied by a set of steps that are necessary to complete the task. 
These steps are typically known \emph{a priori} and thus do not have to be learned from the data.
In fact, for practical applications it is desirable that we could make modifications to this logic without having to discard large parts of the dataset.
The ideal sequences of steps that a dialog would follow to complete the task can be arranged in a graph (see Figure~\ref{fig:doc_schema} for a flow chart that summarizes the graph).
Together with the utterances or actions that are associated with the nodes of this graph, we hence call this a \textit{task schema}, or simply \textit{schema}.
Note, that what we call `schema' is similar to the `task specification' of \cite{bohusRavenClawDialogManagement2009}, but distinct from the `schemas' that only define slots and intents of a task as used by \citet{rastogi2019towards}. 

In a typical supervised model that is trained to, say, predict the next system action for a task-oriented dialog, the schema of the training tasks is implicitly captured by the learned model parameters.
This makes generalizing to a new task difficult, as the implicitly memorized schema will no longer be appropriate \cite{mehri2019structured}. 
With \DATASETNAME\ \textbf{we provide explicit schema representations for each task} and thereby enable models to condition on the schema (see \S\ref{sec:models}).

To collect \DATASETNAME\ we use a Wizard of Oz setup \cite{kelleyIterativeDesignMethodology1984a}, where the system's role is played by a human `wizard'.
Based on our pilot studies, we found that the quality of crowd-sourced dialogs depends strongly on 
\begin{inparaenum}[(i)]
    \item the details of how crowd workers are briefed,
    \item that they have an idea of what they are contributing to with their work,
    \item how varied and interesting they perceive the task to be, and
    \item on the incentive structure for receiving payments.\footnote{See Appendix~\ref{apx:turker_comments} for some feedback from workers.} 
\end{inparaenum}
We refined our approach through extensive internal testing and four rounds of pilot studies. 

Our aim is to create an ecologically valid dataset \cite{de_vries_towards_2020} with the following four attributes, which we believe are crucial for a dataset to be of high quality:
\begin{enumerate}
    \item \textbf{Realistic, variable user behavior}. %
    Realistic dialog rarely strictly follows the ideal path, but is interrupted by small talk, changes of the user's mind, and references to events that happen in the user's environment. %
    In \DATASETNAME\ we capture these behaviors.
    \item \textbf{Progression of difficulty.} %
    We collect three kinds of dialogs: %
    \begin{inparaenum}%
        \item \textit{happy} where the dialog progresses along one of the paths in the schema, %
        \item \textit{unhappy} where the user adds complexity to the task (changing their mind, small talk, etc.), and %
        \item \textit{multi-task} where the user engages in a complex dialog spanning multiple domains and tasks (e.g., restaurant, flight, hotel, ride). %
    \end{inparaenum}%
    The progression of difficulty allows better assessment of dialog models and potential for transfer learning across levels of difficulty.
    \item \textbf{Consistency on the system side}. %
The behavior of a task-oriented dialog system should be largely deterministic and not subject to the whims or personality of the wizard. %
    In particular, we encourage wizards to follow the given task schema(s) as closely as possible.
    \item \textbf{Explicit knowledge base queries}. %
    A large part of developing a dialog system is the implementation of application programming interface (API) calls, such as knowledge base queries. %
    In \DATASETNAME\ we represent our dialogs as a three-party interaction wherein the system acts as the intermediary between a user and a knowledge base (user $\leftrightarrow$ system $\leftrightarrow$ knowledge base). %
    Thus, models have to learn \textit{when} to query the knowledge base, \textit{what} the query should be, and \textit{how} to explain the returned knowledge base item to the user. 
\end{enumerate}

With this paper, we contribute 
\begin{inparaenum}[(i)]
    \item a novel, high-quality dataset of 127,833 utterances and knowledge base queries across 5,820 task-oriented dialogs in 13 domains that admits the above-mentioned properties (with caveats, see \S\ref{sec:corpus}), 
    \item a novel, scalable crowd-sourcing setup to collect arbitrarily large datasets of the same quality as \DATASETNAME, and
    \item novel schema guided dialog models that use the explicit task schemas we provide to generalize to unseen tasks.
\end{inparaenum}
The code for the latter setup, all collected (anonymised) data, and all modeling code is freely available under \anonymous{\DATASETURL}.

\section{Related Work}
\label{sec:related_work}


\begin{table*}[!htb]
\centering
\small
\begin{tabular}{|l|cccccc|}
\hline
& \textbf{MultiWoZ} & \textbf{MetaLWoZ} & \textbf{Frames} & \textbf{Taskmaster} & \textbf{SGD} & \textbf{\DATASETNAME} \\\hline

Number of dialogs & 8,438 & 37,884 & 1,369 & 13,215 & 16,142 & 5,820 \\
Number of domains & 7 & 47 & 1 & 6 & 16 & 13 \\
Number of tasks & 7 & 227 & 1 & 6 & 16 & 24 \\
Total number of turns & 115,434 & 432,036 & 19,986 & 274,647 & 329,964 & 127,833 \\
Average turns / dialog & 13.46 & 11.40 & 14.60 & 21.99 & 20.44 & 21.71 \\
Average tokens / turn & 13.13 & 8.47 & 12.60 & 8.62 & 9.75 & 11.20 \\\hline
Ensure system consistency & \xmark & \xmark & \xmark & \xmark & \xmark & \cmark \\
Real-time data collection & \xmark & \cmark & \cmark & \cmark\footnotemark & \cmark & \cmark \\
Difficulty progression & \cmark & \xmark & \xmark & \xmark & \xmark & \cmark \\
Explicit KB queries & \cmark & \xmark & \cmark & \xmark & \cmark & \cmark \\
\hline
\end{tabular}
\caption{Comparison of dataset statistics and characteristics.}
\label{tab:dataset_statistics}
\end{table*}

\subsection{Dialog Datasets}
\label{sec:related_work:datasets}

While \DATASETNAME\ shares some aspects with existing datasets, to our knowledge it is the first that admits all of the properties listed in the Introduction (see Table~\ref{tab:dataset_statistics} for an overview).
For example, similar to the dataset of the Fourth Dialogue State Tracking Challenge (DSTC4, \citet{kimFourthDialogState2017}) and its successors, \DATASETNAME\ is composed of human-human dialog, yet provides much richer annotation.
Furthermore, similar to the Microsoft Frames dataset \citep{elasriFramesCorpusAdding2017} we make knowledge base queries explicit. 
\DATASETNAME, however, covers 13 domains and encourages a consistent system behavior, while Frames does not do so and only covers 2 domains.
A much larger number of 47 domains is covered by the MetaLWOz dataset \cite{schulzMetaLWOzDatasetMultiDomain2019} which, however, does not provide labels for the system's actions, nor explicit knowledge base queries as \DATASETNAME\ does.
Finally, the data collection procedure for \DATASETNAME\ is scalable and reproducible, so larger datasets can be collected if necessary. 

\footnotetext{%
Only half of the Taskmaster-1 dataset has been collected via real-time human-human conversations. %
The other half --- ``Self-Dialogs" --- have been completely written by a single crowd worker, who is assuming the user and wizard role, each.%
}

In contrast to \DATASETNAME, the MultiWOZ \cite{budzianowskiMultiWOZLargeScaleMultiDomain2018, ericMultiWOZMultiDomainDialogue2019} and Google Taskmaster-1 \cite{byrne2019taskmaster} datasets do not have annotated knowledge base queries (i.e.\ when should the system send what query). 
Furthermore, these datasets suffer from other issues such as weak history dependence and inconsistent system responses \cite{mosigWhereContextCritique2020}.
With \DATASETNAME\ we aim to solve these issues with much more detailed instructions for the wizards and creativity-encouraging prompts for users, as we explain in \S\ref{sec:collection}. 
For example, we occasionally inspire users to assume a certain personality, as in \citet{zhangPersonalizingDialogueAgents2018}.
In addition, we provide response suggestions for the wizards, as was done for the non task oriented BlendedSkillTalk dataset \citep{smithCanYouPut2020}.


Another related and recent dataset is the Schema Guided Dialogue Dataset (SGD), which consists of multi-domain dialogs across 16 domains and also includes meta information about the different domains such as lists of valid slots and intents \cite{rastogi2019towards}. 
Note, that while \citet{rastogi2019towards} call this meta information `schema', it is distinct from our schema that contains information about the ideal dialog flow for each task.

\subsection{Structured Dialog Modeling}

The value of incorporating explicit structure into neural models of dialog has been well recognized. 
RavenClaw \citep{bohusRavenClawDialogManagement2009} disentangles the task specification and the dialog engine for rule-based task-oriented dialog systems. 
\DATASETNAME\ shares a similar motivation and aims to extend to explicitly disentangle the task schema in neural, data-driven models. 
Hybrid Code Networks \cite{williamsHybridCodeNetworks2017} incorporate task-specific constraints to avoid illogical system actions (e.g., requesting known slots) in neural models. 
Recently, several attempts have been made to use intermediate annotations (e.g., belief state, dialog acts) to explicitly incorporate structure in neural dialog models \citep{mehri2019structured,chen2019semantically,peng2020soloist}. 
Furthermore, the SGD dataset by \citet{rastogi2019towards} mentioned in \S\ref{sec:related_work:datasets} comes with explicit slot and intent annotations that serve as an inductive bias for their model.

\section{Data Collection Method}
\label{sec:collection}

We collect our annotated dialogs from Mechanical Turk~\cite{crowston2012amazon}, using ParlAI~\cite{miller2017parlai}.

\subsection{Stages}

We run our data collection in 4 stages to prepare and evaluate workers effectively, and to maximize the quality of the dialogs.
In Stage~I and Stage~III, workers are asked to watch video tutorials and answer questionnaires to qualify for Stage~II and Stage~IV, respectively.
In Stage~II and Stage~IV, workers are paired up to produce single-task and multi-task dialogs, respectively.
The estimated earnings and bonuses increase with each stage.
Workers are made aware of this fact during Stage~I to encourage them to go through all four stages.\footnote{For the worker's reward payments and bonuses, please refer to Appendix~\ref{apx:payment}.}

To enter Stage~I, we require workers to have a minimal approval rate of 98\%, a minimum of $10^4$ approved assignments, and a US-based Mechanical Turk account.
To succeed in the questionnaire (see Appendix~\ref{apx:multiple_choice:single}), workers have to answer 10 questions about the video\footnote{Available from: \anonymous{\url{https://youtu.be/L7QpscLPTFM}}} with no more than 6 hints.
Of the 210 workers that attempted the Stage~I tutorial, 162 (77\%) succeeded.

In every assignment of Stage~II, workers are paired up as ``users'' and ``AI Assistants'' (wizards), and assigned a particular \textit{scenario}. 
A scenario consists of instructions for the user (see \S\ref{sec:collection:user_interface}), instructions for the wizard, and a specialized knowledge base interface for the wizard to handle the particular task(s) at hand (see \S\ref{sec:collection:wizard_interface}).
All scenarios in Stage~II concern one task at a time, e.g.\ only organizing a party or only making a doctor appointment (see \S\ref{sec:corpus} for a list of tasks).
Out of the 162 workers who entered Stage~II, 74 continued into Stage~III.

The Stage~III video tutorial\footnote{Available from: \anonymous{\url{https://youtu.be/dd0s2Sqox6g}}} prepares workers for multi-task dialogs in Stage~IV (see Appendix~\ref{apx:multiple_choice:multi} for the questionnaire).
Of the 74 workers who got the qualification to enter Stage~III, 72 attempted to pass it and all of them progressed to the final Stage~IV.

\subsection{User's Interface}
\label{sec:collection:user_interface}

We want to cover a wide range of user behaviors. 
Therefore, we design the user's interface (see Appendix~\ref{apx:tasks}) to encourage the worker to make the dialog more realistic, i.e.\ to occasionally engage in small talk, refer back to earlier parts of the dialog, change their mind, or use negation.

Therefore, during the dialog, users not only receive messages from the AI Assistant (wizard), but also from the ``MTurk System bot'', which instructs them to, e.g., change their mind about something, interrupt the dialog with chitchat, or refer back to an earlier stage of the dialog.

These in-dialog instructions come in three flavors. 
Sometimes, the instruction is direct and vague, such as \textit{``Either tell a joke, or try to engage in some small talk with the AI Assistant''}.
Other times, the instruction is direct and specific, such as \textit{``Change your mind about either the apartment size and/or ask for the balcony to be on the east side instead of the south side''}.
Finally, some instructions describe things that happen in the user's environment, such as \textit{``You get a text from your hotel: The water pipe broke and -- regrettably -- they cannot accommodate you tonight''}.

Most in-dialog instructions are generated from templates that contain multiple placeholders for specifics (e.g.\ the new balcony side) and reasons (e.g.\ the broken water pipe).
Thus, turkers almost never see the same in-dialog instruction set more than once.

In Stage~IV, most instruction sets merely point out the topics that the user should touch on and provide some example names or locations for them to work with, as this leads to a particularly broad range of behaviors (see Appendix~\ref{apx:example_dialogs} for some examples).

\subsection{Wizard's Interface}
\label{sec:collection:wizard_interface}

In contrast to the user, we want the wizard to behave in a consistent and structured manner.
Therefore, we design the wizard's interface such that they are encouraged to use predefined replies whenever possible, and to not rely on any other sources of information than those provided by the interface.
Most importantly, wizards are instructed to always (when possible) follow a flow chart representation of the task's schema (as shown in Figure~\ref{fig:doc_schema}), which we use to condition our models on particular tasks (see \S\ref{sec:models}).

\begin{figure}[tb]
    \centering
    \includegraphics[width=1.00\columnwidth]{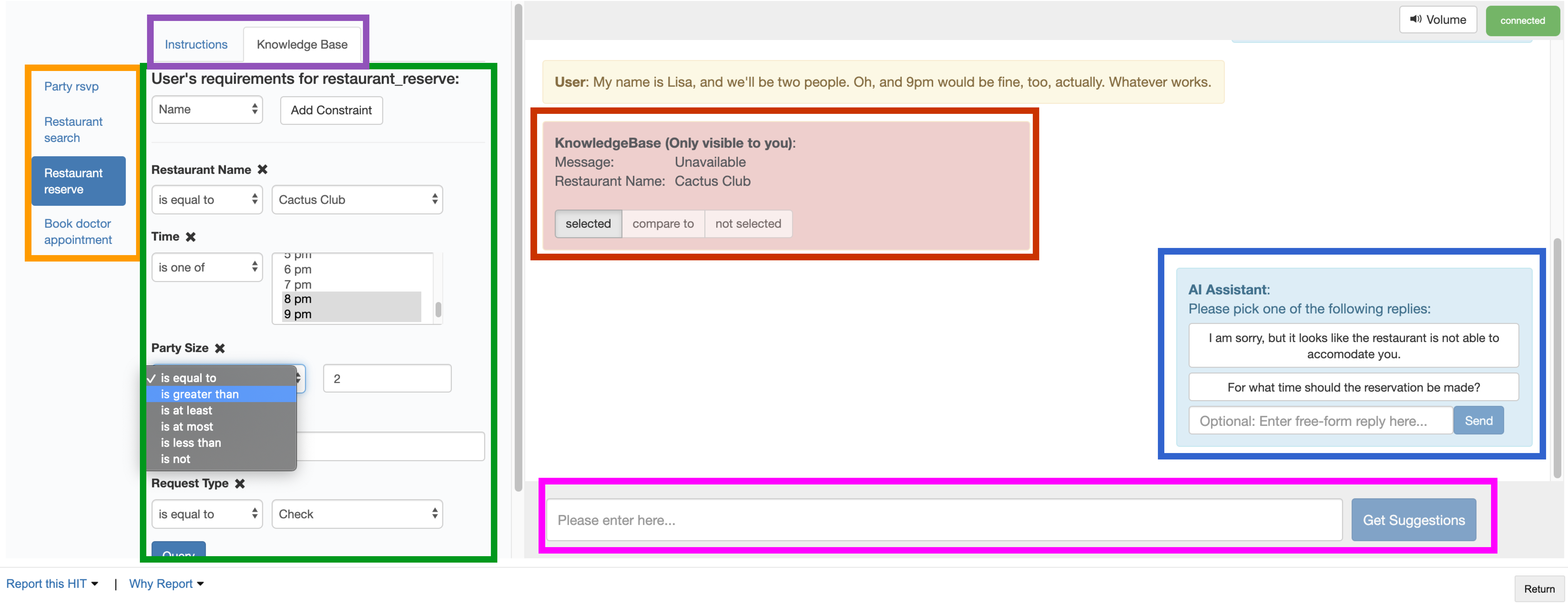}
    \caption{%
      Wizard's graphical user interface with annotation. %
      From left to right, top to bottom. %
      Orange: Tabs to switch tasks. %
      Purple: Tabs to switch between schema flow chart and knowledge base. %
      Green: Knowledge base interface (different for each task).
      Red: Knowledge base item.
      Magenta: Response query field.
      Blue: Suggested responses.
    }
    \label{fig:ui_wizard}
\end{figure}

In the multi-task dialogs (Stage~IV), wizards can switch between up to five tasks by selecting the corresponding tab on the left side of the screen (orange rectangle in Figure~\ref{fig:ui_wizard}). 
We refer to the results of these queries as \textit{knowledge base items} (red rectangle in Figure~\ref{fig:ui_wizard}). 
The knowledge base items are not visible to the user, and thus the wizards have to describe them when required by the schema.

To reply, wizards enter a query into a text box (magenta rectangle in Figure~\ref{fig:ui_wizard}).
They are then presented with a ranked list of suggested responses that they can pick from (blue rectangle). 
If no suggestion fits what they need to say, they can either change the search query or use a custom reply. 
The latter should only be necessary in situations where the user shows unexpected behavior, and thus the schema cannot provide guidance.

The response suggestions work by classifying the intent of the wizard's input, conditional on the selected task and knowledge base item. 
For the classification we use 24 instances of the recently proposed DIET architecture by \citet{bunkDIETLightweightLanguage2020}, one for each task.
Every intent is then linked to a response string template, which in turn is filled with the information from the selected knowledge base item (if any).


From a typical wizard turn we thus collect 
\begin{inparaenum}[(i)]
    \item which task the wizard is considering,
    \item the knowledge base query (optional),
    \item the selected knowledge base item (optional),
    \item the reply text entered, and
    \item the selected response text and action label.
\end{inparaenum}
See Appendix~\ref{apx:format} for details on the data format.

Together, the schema flow charts, knowledge base forms, and response suggestions provide a framework for wizards to make their behavior consistent and structured, even when dealing with erratic users.

\section{The \DATASETNAME\ Dataset}
\label{sec:corpus}

Our dataset covers 24 tasks in the 13 domains, with a total of 127,833 turns in 5,820 dialogs (i.e. 22 turns per dialog on average).
Of these, 4,152 (71.3\%) are single-task dialogs,
and 1,668 (28.7\%) are multi-task dialogs.
Of the 4,152 single task dialogs, 2,688 follow a happy path, i.e.\ the user is not instructed to do anything that is not accounted for by the schema.
A detailed list and description of the tasks, as well as detailed turn statistics can be found in Appendix~\ref{apx:tasks}. 

We focused our data collection design choices around grounding dialogs in knowledge base queries, and providing an explicit mechanism that encourages consistent system actions, which are paramount for real-world conversational assistants.
Furthermore, we required that our dialogs are created in real-time by two humans, and ensured that only workers that are comfortable with our task setup progress to the more advanced multi-task dialogs.

Our goal of high variability of the user's behavior is difficult to quantify, but one indicator that users are more creative than in other datasets is the size of the vocabulary used.
Specifically, users in \DATASETNAME\ employ 6507 distinct dictionary words to express themselves (here we do not count entities). 
This is 3.6 times as large a vocabulary as users in MultiWOZ use.
Even when we take into account that STAR covers almost twice as many domains, this indicates more complexity on the user side.

Evaluating the consistency of the wizard's behavior is even more challenging, because we did not annotate user intents in the dataset, which makes it more difficult to assess whether the next wizard action is consistent given the user's utterance. 
For example, we have observed that when users begin the dialogue with, say, the second piece of information that the wizard should collect according to the schema, wizards often continue asking for the third piece of information even though they should technically ask for the first (if that was not also given by the user).
Other inconsistencies can occur later in the dialogue, but with increasing history length it becomes less and less clear if two dialogue states should be considered the same or not.
We can, however, see if wizards generally ask questions in the right order at the beginning of a single-task dialog. 
Table~\ref{tab:correctly_ordered_fractions} in Appendix~\ref{apx:tasks} shows how many dialogs begin with questions in the expected order.
Here, skipping questions or asking questions twice is allowed, since users might provide more than one (or no) piece of information at a time.
On average, in 91\% of all single-task dialogs the wizards follow the correct order of actions at the beginning of the dialog.
Since this is not the only possible source of inconsistencies, we think that our data collection setup should be improved to enforce system consistency further. 
However, this number also suggests that a large portion of the data \textit{is} consistent and can be used to train supervised learning algorithms.

\begin{figure}[!htb]
    \centering
    \includegraphics[width=1.00\columnwidth]{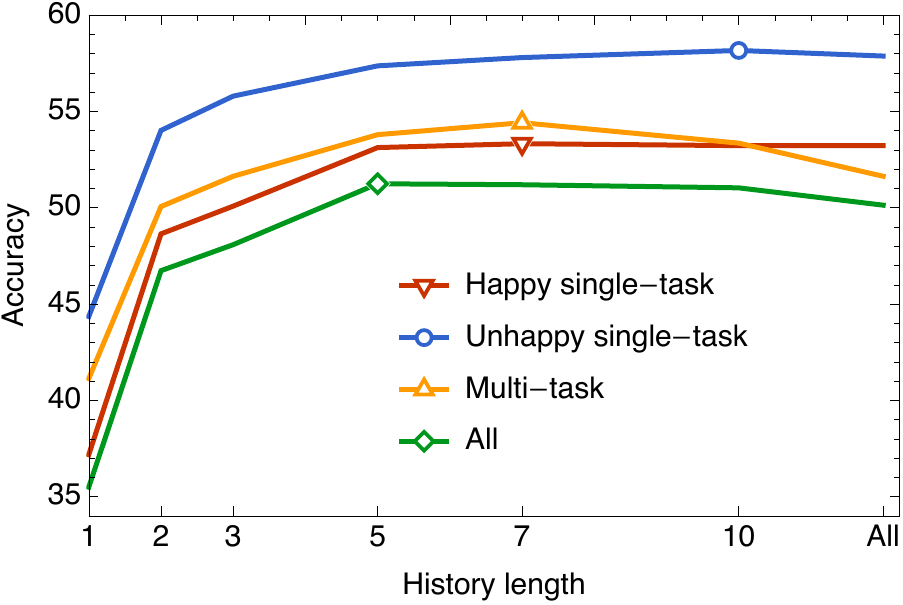}
    \caption{Accuracy of the response selector vs.\ number of turns that it takes into account.}
    \label{fig:history_dependence}
\end{figure}

An essential feature of task-oriented dialog is its history dependence, i.e.\ the next system action hinges on what has been said or decided in \textit{multiple} previous turns, as this sets it apart from question answering settings.
To assess the history dependence of \DATASETNAME, we train a transformer-based response selector~\citep{bunkDIETLightweightLanguage2020} with ConveRT word embeddings \cite{hendersonConveRTEfficientAccurate2019} for predicting the next wizard utterance with a varying number of preceding dialog turns that the model has access to.
The results, presented in Figure~\ref{fig:history_dependence}, indicate that system actions depend on 5 to 10 turns of history, which highlights the complexity of our collected dialogs.

\begin{figure*}[htb]
    \centering
    \includegraphics[width=0.95\textwidth]{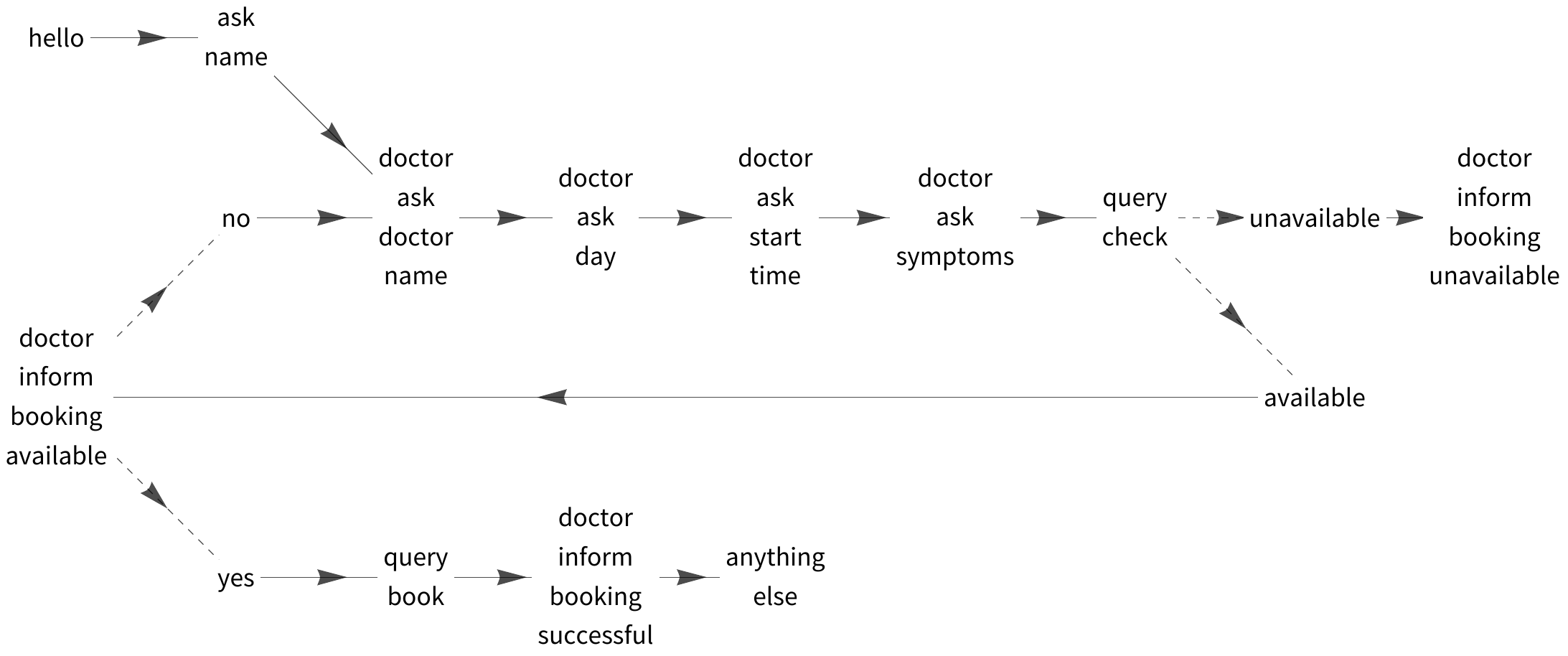}
    \cprotect\caption{%
        Schema graph representation corresponding to the flow chart visualized in Figure~\ref{fig:doc_schema}.
    }
    \label{fig:doc_schema_graph}
\end{figure*}

\section{Models}
\label{sec:models}

\begin{figure*}[ht]
    \centering
    \includegraphics[width=0.9\textwidth]{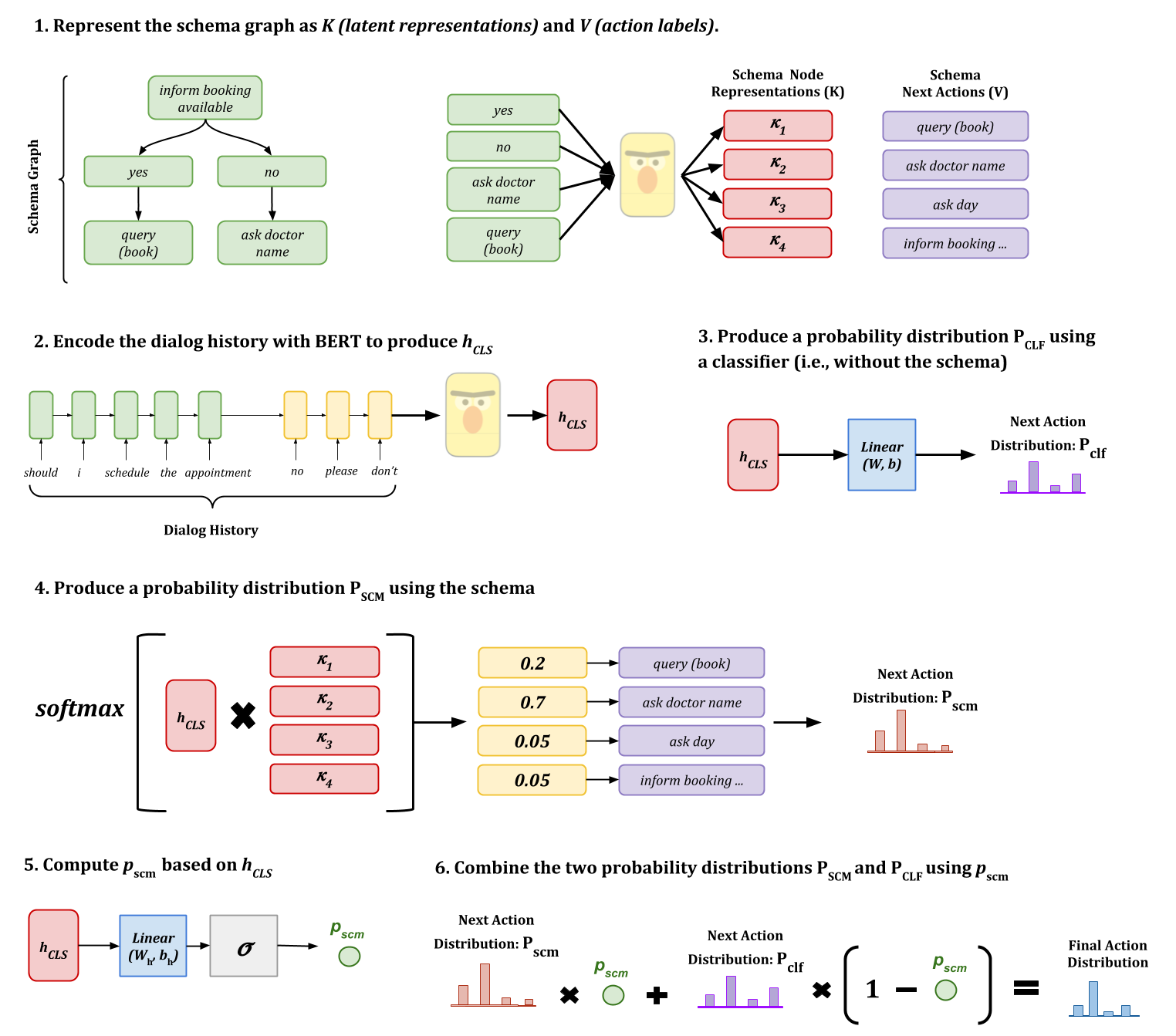}
    \caption{Schema-guided next action prediction model as described in \S\ref{sec:models:action} in Equations 1 - 4.}    
    \label{fig:model}
\end{figure*}

\DATASETNAME~allows us to train models which are conditioned on a task-specific schema. Rather than relying on a model to implicitly learn the steps required to complete a task (e.g., ask for a name, then ask for a phone number), we instead propose to explicitly provide this information through the task-specific schemas. 
We hypothesize that such models will better generalize to tasks that are unseen at training time. 
In this Section, we present baseline models for next action prediction and response generation, both with and without conditioning on the schema. 

We first introduce a general mechanism of representing the schema in \S\ref{sec:models:schema}. 
Next we discuss both schema-free and schema-guided models for the tasks of next action prediction in \S\ref{sec:models:action} and response generation in \S\ref{sec:models:generation}.

\subsection{Schema Representation}
\label{sec:models:schema}

The schema is a graph, wherein each node is associated with text that either describes a system or user action. 
For example, the schema in Figure~\ref{fig:doc_schema_graph} can be constructed from the flow chart of Figure~\ref{fig:doc_schema}. 
Each bullet point in the flow chart becomes a node in the graph, and every node would be associated either with a system response template or with text describing the user's anticipated input (e.g. `No'). 
The complete translation of Figure~\ref{fig:doc_schema} into a schema is given in Appendix~C.

While collecting \DATASETNAME, we designed the instructions and the task-specific flow charts to encourage that the wizard's actions are deterministic. 
As such, if we accurately determine our current position in the schema graph (i.e., the state of the dialog), the next system action can be determined by following the single outgoing edge. 
For example, if we consider a dialog (corresponding to the schema in Figure~\ref{fig:doc_schema_graph}) wherein the system has requested the user's name and the doctor's name -- we know that our current position in the graph is \texttt{doctor-ask-doctor-name} and therefore the next system action should be \texttt{doctor-ask-day}.

We manually construct the schema graphs by considering the flow charts and a few example dialogs. 
While we believe this graph-based representation of the schema to be derived naturally from the formulation of the flow charts, it is by no means the \textit{only} possible representation of the schema. 
We consider the schema representation to be an attribute of the models rather than of the dataset. 
As such, we anticipate that future work will propose improved representations of the schemas, which may involve manually constructing schemas for the 24 tasks or devising an automatic data-driven mechanism of extracting schema representations.


\subsection{Next Action Prediction}
\label{sec:models:action}

We introduce both schema-free and schema-guided BERT baselines for next action prediction on \DATASETNAME. 
For the schema-free model, we begin by encoding the dialog history using the BERT-base model \citep{devlin2018bert}. 
We use the pooled representation, denoted $\mathbf{h}_{CLS}$ as the latent representation of the dialog history. 
This representation is passed through a linear layer (parameterized by $W$ and $b$) to produce a probability distribution over the set of possible actions:
\begin{equation}
    P_{\rm{clf}} = \rm{softmax}(W\,\mathbf{h}_{CLS} + b)\;.
\end{equation}

The schema-guided model augments the BERT classifier by conditioning it on the aforementioned schema graph. The schema-guided models use the schema to produce a probability distribution over the set of actions, $P_{\rm{scm}}$. This probability distribution is then combined with $P_{\rm{clf}}$ to produce the final distribution over the set of actions. To produce $P_{\rm{scm}}$ we begin by encoding the dialog history, $\mathbf{h}_{CLS}$. Next, we produce a latent representation for each node in the schema by using BERT to encode the associated text. We denote the latent representations of the nodes as $K = \{\kappa_1, \kappa_2, \dots, \kappa_n\}$, such that $\kappa_i$ is the latent representation of the $i$-th node. For each node, we consider the corresponding \textit{next action} to be the action label of the subsequent node (e.g., in Figure \ref{fig:doc_schema_graph}, the \textit{next action} for the \texttt{hello} node would be \texttt{ask-name}). We represent the set of \textit{next actions} as $V = \{v_1, v_2, \dots, v_n \}$ such that $v_i$ is the one-hot representation of the next action corresponding to the $i$-th node. Given $\mathbf{h}_{CLS}$, $K$ and $V$, we compute:

\begin{equation}
P_{\rm{scm}} = \rm{softmax}(\mathbf{h}_{CLS}^T \,K) \cdot V\;
\end{equation}

\DATASETNAME\ includes dialogs where the user exhibits behavior that forces the wizard to deviate from the task schema. To account for this, $P_{\rm{scm}}$ is combined with $P_{\rm{clf}}$, by computing $\rho_{\rm{scm}}$ and using it to combine the two probability distributions. We compute $\rho_{\rm{scm}}$ by passing $\mathbf{h}_{CLS}$ through a linear layer (defined by $W_h$ and $b_h$):
\begin{align}
    \rho_{\rm{scm}} &= \sigma(W_{h}^T\,\mathbf{h}_{CLS} + b_h) \;, \\
    P_{\rm{fin}} &= \rho_{\rm{scm}}\,P_{\rm{scm}} + (1 - \rho_{\rm{scm}})\,P_{\rm{clf}}\;
\end{align}
where $\sigma$ denotes the logistic sigmoid function.

\subsection{Generation}
\label{sec:models:generation}

We fine-tune GPT-2 \citep{radford2019language} for the task of response generation on \DATASETNAME. 
We carry out response generation experiments both with and without the schema. In the schema-free model, we fine-tune GPT-2 to produce the response conditioned on the dialog history. 
During inference, we sequentially decode until the end of sentence token is generated.

We augment the GPT-2 model to condition on the schema representation. 
To do so, we use the aforementioned schema-guided BERT classifier (pictured in Figure~\ref{fig:model}) to produce a probability distribution over the set of next actions. 
We then use the top 3 predicted next actions\footnote{We determined that using the top 3 actions results in best performance and run-time, through preliminary experiments.} to produce a schema-augmented dialog context.
Concretely, given a dialog history $H$, we use the schema-guided next action classifier to predict the top 3 actions, $a_1$, $a_2$ and $a_3$. 
For each of these actions, we identify the corresponding system response template, $t_1$, $t_2$ and $t_3$. 
We then concatenate the history and the response templates to produce the schema-augmented dialog context: $H ; t_1 ; t_2  ; t_3 $, where $;$ denotes a separator token. 
GPT-2 is then fine-tuned to generate the ground-truth response conditioned on this schema-augmented dialog context.


The schema-guided next action classifier uses the explicit schema representation to predict the next actions and our schema-guided response generation model uses the templates corresponding to the predicted next actions. 
As such, our schema-guided response generation model explicitly uses the schema graph to determine the next system action. 
We hypothesize that this will result in better performance, particularly when transferring to unseen tasks or domains.

\section{Experiments}
\label{sec:experiments}

Here we demonstrate how to use the \DATASETNAME\ dataset for the tasks of next action prediction and response generation. Furthermore, we carry out zero-shot transfer learning experiments to assess our models' performance on unseen tasks and domains. 

Unless otherwise specified, we carried out the following experiments in three stages: 
\begin{inparaenum}[(i)]
    \item happy,
    \item unhappy, and
    \item multi-task.
\end{inparaenum} 
For each stage, we trained the models on 80\% of the data from the current stage and all data from the previous stages. 
We tested models on the remaining 20\% of data from the current stage. 
We forego the use of a validation set. 
If necessary, future work should either use cross-validation or use a small amount of the training data for validation. 
Only for the zero-shot experiments we segment the data based on task or domain.

\begin{table}[ht]
    \centering
    \small
    \begin{tabular}{|l|c|c|c|}
    \hline
        Model & Happy & Unhappy & Multi-task  \\
         & F-1 & F-1 & F-1  \\ \hline
        BERT &  \textbf{73.30} & \textbf{73.93} & \textbf{73.61} \\
        BERT + \textit{schema} & 71.09 & 72.28 & 73.13  \\ \hline
    \end{tabular}
    \caption{Results of the next action prediction experiments across three stages. Note that the results are not comparable across the three stages, since the latter stages were trained on more data.}
    \label{tab:intent_results}
\end{table}

\begin{table*}[ht]
    \centering
    \small
    \begin{tabular}{|c|c|c|c|c|c|c|c|c|c|}
    \hline
        Model & \multicolumn{3}{|c|}{Happy} & \multicolumn{3}{|c|}{Unhappy} & \multicolumn{3}{|c|}{Multi-task}  \\ \hline
         & IEM & BLEU & Entity F-1 & IEM & BLEU & Entity F-1  & IEM & BLEU & Entity F-1   \\ \hline
        GPT-2 & 42.93 & 58.60 & 75.85 & 46.63 & \textbf{61.44} & 78.21 & 35.18 & 47.75 & 60.79 \\
        GPT-2 + \textit{schema} & \textbf{44.17} & \textbf{58.96} & \textbf{77.89} & \textbf{47.03} & 60.39 & \textbf{79.45} & \textbf{50.43} & \textbf{61.35} & \textbf{76.42}\\ \hline
    \end{tabular}
    \normalsize
    \caption{Results of the generation prediction experiments for the three stages. Note that the results are not comparable across the three stages, since the latter stages were trained on more data.}
    \label{tab:gen_results}
\end{table*}

\subsection{Next Action Prediction}

In next action prediction, the task is to predict the next system action conditioned on the dialog history. 
For every response, we provide a single ground-truth action. We evaluate our models on this task using the weighted F-1 score. 
The results shown in Table~\ref{tab:intent_results} demonstrate that the BERT classifier outperforms the schema-augmented BERT model.
While this is a negative result, it should be noted that the schema is intended to facilitate transfer learning and the fact that it does not yield performance improvements when evaluating on \textit{seen} tasks and domains does not invalidate its effectiveness.

\subsection{Response Generation}

Response generation is the task of generating the system response conditioned on the dialog history. 
Through the schema and the suggested response templates, \DATASETNAME\ was designed to contain consistent and deterministic system behavior. 
As a result, it is suitable to evaluate with word-overlap metrics since the one-to-many problem \citep{zhao2017learning} is less prevalent. 
We evaluate response generation with three metrics: 
\begin{inparaenum}[(i)]
    \item BLEU-4 \citep{papineni2002bleu},
    \item IEM (in-domain exact match), which measures the rate of exact matches between the hypothesis and domain-specific reference responses (i.e., excluding 'Hello', 'Goodbye', etc.), and
    \item Entity F-1 \citep{wen2016network} which measures the weighted F-1 for entities in the response (e.g., `2 pm').
\end{inparaenum}

The results of the generation experiments are shown in Table~\ref{tab:gen_results}. 
Leveraging the schema leads to a performance increase, albeit less so for the unhappy dialogs, as one would expect (users in unhappy dialogs diverge from the schema). 
Leveraging the schema results in a significant performance increase in the multi-task setting, suggesting that explicitly conditioning on the schema (both through the next action classifier and through the response templates) is valuable for multi-task dialogs.

\subsection{Other Tasks}
\DATASETNAME\ can be used for additional tasks, such as \textbf{knowledge base query prediction} (predicting the correct knowledge base keys, values and operators), \textbf{schema prediction} (predicting the schema of a task, given a collection of dialogs), and \textbf{out-of-domain detection} (detecting whether a user has made an out-of-domain request).


\subsection{Zero-Shot Transfer}

The variety of tasks and domains as well as the structured policy representation makes the \DATASETNAME\ dataset ideal for zero-shot transfer learning experiments. 
For the tasks of next action prediction and response generation, we carry out two sets of transfer experiments: (i) using only happy dialogs and (ii) using both happy and unhappy dialogs. 
We also experiment with both \textit{task} transfer and \textit{domain} transfer. 
While there might be high overlap between tasks (e.g., \textit{bank\_balance} and \textit{bank\_fraud\_report}) there is less overlap between domains (e.g., \textit{hotel} and \textit{flight}). 

We train the model on $N-1$ tasks/domains and evaluate on the $N$-th. 
We repeat this for each task/domain for a total of $N$ times ($N = 24$ for task transfer, $N=13$ for domain transfer). 

\begin{table}[ht]
    \centering
    \small
    \begin{tabular}{|c|c|c|c|}
    \hline
        Experiment & Model & \textit{H} & \textit{H+U}   \\
                & & F-1 & F-1 \\ \hline
        \multirow{2}{*}{task transfer}  & BERT &  36.45 & 36.89   \\
             & BERT + \textit{schema} & \textbf{36.77} & \textbf{37.15} \\ \hline
        \multirow{2}{*}{domain transfer}  & BERT &  34.84 & 35.63   \\
               & BERT + \textit{schema} & \textbf{37.20} & \textbf{35.71} \\ \hline
    \end{tabular}
    \normalsize
    \caption{Results of the zero-shot \textit{task} and \textit{domain} transfer experiments for next action prediction. The column denoted \textit{H} refers to the experiments with only the happy dialogs, while \textit{H+U} denotes the use of both the happy and unhappy dialogs.}
    \label{tab:task_domain_intent_results}
\end{table}
%
         
%
\begin{table}[ht]
    \centering
    \small
    \begin{tabular}{|c|c|c|c|}
    \hline
        \multicolumn{4}{|c|}{Happy}  \\ \hline
        Model & IEM & BLEU & Entity F-1     \\ \hline
        GPT-2 & 4.76 & 15.23 & 48.22  \\
        GPT-2 + \textit{schema} & \textbf{5.17} & \textbf{15.49} & \textbf{52.12}  \\ \hline
        \multicolumn{4}{|c|}{Happy + Unhappy}  \\ \hline
        Model & IEM & BLEU & Entity F-1     \\ \hline
        GPT-2 & 8.61 & 14.57 & 47.45  \\
        GPT-2 + \textit{schema} & \textbf{8.79} & \textbf{15.15} & \textbf{51.63}  \\ \hline
    \end{tabular}
    \normalsize
    \caption{Results of the zero-shot \textit{task} transfer experiments for response generation.}
    \label{tab:task_gen_results}
\end{table}
\begin{table}[ht]
    \centering
    \small
    \begin{tabular}{|c|c|c|c|}
    \hline
        \multicolumn{4}{|c|}{Happy}  \\ \hline
        Model & IEM & BLEU & Entity F-1     \\ \hline
        GPT-2 & 8.14 & 17.12 & 48.18  \\
        GPT-2 + \textit{schema} & \textbf{8.39} & \textbf{18.60} & \textbf{50.23}  \\ \hline
        \multicolumn{4}{|c|}{Happy + Unhappy}  \\ \hline
        Model & IEM & BLEU & Entity F-1     \\ \hline
        GPT-2 & 8.77 & 19.46 & 50.43  \\
        GPT-2 + \textit{schema} & \textbf{8.82} & \textbf{19.74} & \textbf{53.02}  \\ \hline
    \end{tabular}
    \normalsize
    \caption{Results of the zero-shot \textit{domain} transfer experiments for response generation.}
    \label{tab:domain_gen_results}
\end{table}

The results in Tables~\ref{tab:task_domain_intent_results}-\ref{tab:domain_gen_results} show that the use of the schema consistently results in performance improvement for zero-shot task and domain transfer. 
These preliminary results highlight the potential of using the schema as an inductive bias to help with model generalizability and task transfer. 
However, the zero-shot transfer experiments demonstrate that even the schema-guided models perform significantly worse on \textit{unseen} tasks than it does on seen tasks. To mitigate this performance gap, future work should explore mechanisms of better leveraging the task-specific schemas to facilitate generalizability to unseen tasks and domains.

\section{Conclusions}
\label{sec:conclusions}

With this work, we make multiple contributions to the field of task-oriented dialog research.
First, we presented \DATASETNAME, a novel dialog dataset that we specifically designed to facilitate transfer learning experiments.
Second, we introduced a new, scalable crowd sourcing paradigm to collect data of similar quality as \DATASETNAME.
In future work, this setup could be used to expand on \DATASETNAME\ by collecting data for additional tasks, domains, or in languages other than English.
Finally, we established baseline scores for next action prediction, response generation, and zero-shot transfer learning for the former two tasks.
With this we demonstrated how task schemas can be used to improve transfer learning capabilities.
We also outlined a variety of other experiments that \DATASETNAME\ would be suitable for, and we look forward to seeing these experiments, as well as improvements upon our baseline scores, implemented in future publications.

\clearpage
\newpage


%
\bibliography{transferdialogue}

\begin{thebibliography}{32}
\expandafter\ifx\csname natexlab\endcsname\relax\def\natexlab#1{#1}\fi

\bibitem[{Adiwardana et~al.(2020)Adiwardana, Luong, So, Hall, Fiedel,
  Thoppilan, Yang, Kulshreshtha, Nemade, Lu et~al.}]{adiwardana2020towards}
Daniel Adiwardana, Minh-Thang Luong, David~R So, Jamie Hall, Noah Fiedel, Romal
  Thoppilan, Zi~Yang, Apoorv Kulshreshtha, Gaurav Nemade, Yifeng Lu, et~al.
  2020.
\newblock Towards a human-like open-domain chatbot.
\newblock \emph{arXiv preprint arXiv:2001.09977}.

\bibitem[{Bohus and Rudnicky(2009)}]{bohusRavenClawDialogManagement2009}
Dan Bohus and Alexander~I. Rudnicky. 2009.
\newblock \href {https://doi.org/10.1016/j.csl.2008.10.001} {The {{RavenClaw}}
  dialog management framework: {{Architecture}} and systems}.
\newblock \emph{Computer Speech \& Language}, 23(3):332--361.

\bibitem[{Budzianowski et~al.(2018)Budzianowski, Wen, Tseng, Casanueva, Ultes,
  Ramadan, and Ga{\v{s}}i{\'c}}]{budzianowskiMultiWOZLargeScaleMultiDomain2018}
Pawe{\l} Budzianowski, Tsung-Hsien Wen, Bo-Hsiang Tseng, I{\~n}igo Casanueva,
  Stefan Ultes, Osman Ramadan, and Milica Ga{\v{s}}i{\'c}. 2018.
\newblock \href {https://doi.org/10.18653/v1/D18-1547} {{M}ulti{WOZ} - a
  large-scale multi-domain {W}izard-of-{O}z dataset for task-oriented dialogue
  modelling}.
\newblock In \emph{Proceedings of the 2018 Conference on Empirical Methods in
  Natural Language Processing}, pages 5016--5026, Brussels, Belgium.
  Association for Computational Linguistics.

\bibitem[{{Bunk} et~al.(2020){Bunk}, {Varshneya}, {Vlasov}, and
  {Nichol}}]{bunkDIETLightweightLanguage2020}
Tanja {Bunk}, Daksh {Varshneya}, Vladimir {Vlasov}, and Alan {Nichol}. 2020.
\newblock \href {http://arxiv.org/abs/2004.09936} {{DIET: Lightweight Language
  Understanding for Dialogue Systems}}.
\newblock \emph{arXiv e-prints}, page arXiv:2004.09936.

\bibitem[{Byrne et~al.(2019)Byrne, Krishnamoorthi, Sankar, Neelakantan,
  Goodrich, Duckworth, Yavuz, Dubey, Kim, and Cedilnik}]{byrne2019taskmaster}
Bill Byrne, Karthik Krishnamoorthi, Chinnadhurai Sankar, Arvind Neelakantan,
  Ben Goodrich, Daniel Duckworth, Semih Yavuz, Amit Dubey, Kyu-Young Kim, and
  Andy Cedilnik. 2019.
\newblock Taskmaster-1: Toward a realistic and diverse dialog dataset.
\newblock In \emph{Proceedings of the 2019 Conference on Empirical Methods in
  Natural Language Processing and the 9th International Joint Conference on
  Natural Language Processing (EMNLP-IJCNLP)}, pages 4506--4517.

\bibitem[{Chen et~al.(2019)Chen, Chen, Qin, Yan, and
  Wang}]{chen2019semantically}
Wenhu Chen, Jianshu Chen, Pengda Qin, Xifeng Yan, and William~Yang Wang. 2019.
\newblock Semantically conditioned dialog response generation via hierarchical
  disentangled self-attention.
\newblock \emph{arXiv preprint arXiv:1905.12866}.

\bibitem[{Crowston(2012)}]{crowston2012amazon}
Kevin Crowston. 2012.
\newblock Amazon mechanical turk: {{A}} research tool for organizations and
  information systems scholars.
\newblock In \emph{Shaping the Future of {{ICT}} Research. {{Methods}} and
  Approaches}, pages 210--221. {Springer Berlin Heidelberg}.

\bibitem[{Devlin et~al.(2018)Devlin, Chang, Lee, and
  Toutanova}]{devlin2018bert}
Jacob Devlin, Ming-Wei Chang, Kenton Lee, and Kristina Toutanova. 2018.
\newblock Bert: Pre-training of deep bidirectional transformers for language
  understanding.
\newblock \emph{arXiv preprint arXiv:1810.04805}.

\bibitem[{El~Asri et~al.(2017)El~Asri, Schulz, Sharma, Zumer, Harris, Fine,
  Mehrotra, and Suleman}]{elasriFramesCorpusAdding2017}
Layla El~Asri, Hannes Schulz, Shikhar Sharma, Jeremie Zumer, Justin Harris,
  Emery Fine, Rahul Mehrotra, and Kaheer Suleman. 2017.
\newblock \href {https://doi.org/10.18653/v1/W17-5526} {Frames: A corpus for
  adding memory to goal-oriented dialogue systems}.
\newblock In \emph{Proceedings of the 18th {{Annual SIGdial Meeting}} on
  {{Discourse}} and {{Dialogue}}}, pages 207--219. {Association for
  Computational Linguistics}.

\bibitem[{Eric et~al.(2019)Eric, Goel, Paul, Sethi, Agarwal, Gao, and
  Hakkani-Tür}]{ericMultiWOZMultiDomainDialogue2019}
Mihail Eric, Rahul Goel, Shachi Paul, Abhishek Sethi, Sanchit Agarwal, Shuyang
  Gao, and Dilek Hakkani-Tür. 2019.
\newblock \href {http://arxiv.org/abs/1907.01669} {Multiwoz 2.1: Multi-domain
  dialogue state corrections and state tracking baselines}.
\newblock \emph{CoRR}, abs/1907.01669.

\bibitem[{Gao et~al.(2018)Gao, Galley, and Li}]{gao2018neural}
Jianfeng Gao, Michel Galley, and Lihong Li. 2018.
\newblock Neural approaches to conversational ai.
\newblock In \emph{The 41st International ACM SIGIR Conference on Research \&
  Development in Information Retrieval}, pages 1371--1374.

\bibitem[{Henderson et~al.(2019)Henderson, Casanueva, Mrk{\v{s}}i{\'c}, Su,
  Wen, and Vuli{\'c}}]{hendersonConveRTEfficientAccurate2019}
Matthew Henderson, I{\~n}igo Casanueva, Nikola Mrk{\v{s}}i{\'c}, Pei-Hao Su,
  Tsung-Hsien Wen, and Ivan Vuli{\'c}. 2019.
\newblock Convert: Efficient and accurate conversational representations from
  transformers.
\newblock \emph{arXiv}, pages arXiv--1911.

\bibitem[{Kelley(1984)}]{kelleyIterativeDesignMethodology1984a}
John~F Kelley. 1984.
\newblock An iterative design methodology for user-friendly natural language
  office information applications.
\newblock \emph{ACM Transactions on Information Systems (TOIS)}, 2(1):26--41.

\bibitem[{Kim et~al.(2017)Kim, D’Haro, Banchs, Williams, and
  Henderson}]{kimFourthDialogState2017}
Seokhwan Kim, Luis~Fernando D’Haro, Rafael~E Banchs, Jason~D Williams, and
  Matthew Henderson. 2017.
\newblock The fourth dialog state tracking challenge.
\newblock In \emph{Dialogues with Social Robots}, pages 435--449. Springer.

\bibitem[{Kim et~al.(2019)Kim, Galley, Gunasekara, Lee, Atkinson, Peng, Schulz,
  Gao, Li, Adada, Huang, Lastras, Kummerfeld, Lasecki, Hori, Cherian, Marks,
  Rastogi, Zang, Sunkara, and Gupta}]{DSTC8}
Seokhwan Kim, Michel Galley, Chulaka Gunasekara, Sungjin Lee, Adam Atkinson,
  Baolin Peng, Hannes Schulz, Jianfeng Gao, Jinchao Li, Mahmoud Adada, Minlie
  Huang, Luis Lastras, Jonathan~K. Kummerfeld, Walter~S. Lasecki, Chiori Hori,
  Anoop Cherian, Tim~K. Marks, Abhinav Rastogi, Xiaoxue Zang, Srinivas Sunkara,
  and Raghav Gupta. 2019.
\newblock The eighth dialog system technology challenge.
\newblock \emph{arXiv preprint}.

\bibitem[{Mehri et~al.(2019)Mehri, Srinivasan, and
  Eskenazi}]{mehri2019structured}
Shikib Mehri, Tejas Srinivasan, and Maxine Eskenazi. 2019.
\newblock Structured fusion networks for dialog.
\newblock \emph{arXiv preprint arXiv:1907.10016}.

\bibitem[{{Miller} et~al.(2017){Miller}, {Feng}, {Fisch}, {Lu}, {Batra},
  {Bordes}, {Parikh}, and {Weston}}]{miller2017parlai}
A.~H. {Miller}, W.~{Feng}, A.~{Fisch}, J.~{Lu}, D.~{Batra}, A.~{Bordes},
  D.~{Parikh}, and J.~{Weston}. 2017.
\newblock Parlai: A dialog research software platform.
\newblock \emph{arXiv preprint arXiv:{1705.06476}}.

\bibitem[{Mosig et~al.(2020)Mosig, Vlasov, and
  Nichol}]{mosigWhereContextCritique2020}
Johannes~EM Mosig, Vladimir Vlasov, and Alan Nichol. 2020.
\newblock Where is the context?--a critique of recent dialogue datasets.
\newblock \emph{arXiv preprint arXiv:2004.10473}.

\bibitem[{Papineni et~al.(2002)Papineni, Roukos, Ward, and
  Zhu}]{papineni2002bleu}
Kishore Papineni, Salim Roukos, Todd Ward, and Wei-Jing Zhu. 2002.
\newblock Bleu: a method for automatic evaluation of machine translation.
\newblock In \emph{Proceedings of 40th Annual Meeting of the Association for
  Computational Linguistics}, pages 311--318. Association for Computational
  Linguistics.

\bibitem[{Peng et~al.(2020)Peng, Li, Li, Shayandeh, Liden, and
  Gao}]{peng2020soloist}
Baolin Peng, Chunyuan Li, Jinchao Li, Shahin Shayandeh, Lars Liden, and
  Jianfeng Gao. 2020.
\newblock Soloist: Few-shot task-oriented dialog with a single pre-trained
  auto-regressive model.
\newblock \emph{arXiv preprint arXiv:2005.05298}.

\bibitem[{Radford et~al.(2019)Radford, Wu, Child, Luan, Amodei, and
  Sutskever}]{radford2019language}
Alec Radford, Jeff Wu, Rewon Child, David Luan, Dario Amodei, and Ilya
  Sutskever. 2019.
\newblock Language models are unsupervised multitask learners.
\newblock \emph{OpenAI blog}.

\bibitem[{Rastogi et~al.(2019)Rastogi, Zang, Sunkara, Gupta, and
  Khaitan}]{rastogi2019towards}
Abhinav Rastogi, Xiaoxue Zang, Srinivas Sunkara, Raghav Gupta, and Pranav
  Khaitan. 2019.
\newblock Towards scalable multi-domain conversational agents: The
  schema-guided dialogue dataset.
\newblock \emph{arXiv preprint arXiv:1909.05855}.

\bibitem[{Roller et~al.(2020)Roller, Dinan, Goyal, Ju, Williamson, Liu, Xu,
  Ott, Shuster, Smith et~al.}]{roller2020recipes}
Stephen Roller, Emily Dinan, Naman Goyal, Da~Ju, Mary Williamson, Yinhan Liu,
  Jing Xu, Myle Ott, Kurt Shuster, Eric~M Smith, et~al. 2020.
\newblock Recipes for building an open-domain chatbot.
\newblock \emph{arXiv preprint arXiv:2004.13637}.

\bibitem[{Schulz et~al.(2019)Schulz, Atkinson, Adada, Suleman, and
  Sharma}]{schulzMetaLWOzDatasetMultiDomain2019}
Hannes Schulz, Adam Atkinson, Mahmoud Adada, Kaheer Suleman, and Shikhar
  Sharma. 2019.
\newblock \href {https://www.microsoft.com/en-us/research/project/metalwoz/}
  {{{MetaLWOz}}: {{A Dataset}} of {{Multi}}-{{Domain Dialogues}} for the {{Fast
  Adaptation}} of {{Conversation Models}}}.
\newblock \emph{Microsoft Research}.

\bibitem[{Smith et~al.(2020)Smith, Williamson, Shuster, Weston, and
  Boureau}]{smithCanYouPut2020}
Eric~Michael Smith, Mary Williamson, Kurt Shuster, Jason Weston, and Y-Lan
  Boureau. 2020.
\newblock Can you put it all together: Evaluating conversational agents'
  ability to blend skills.
\newblock \emph{arXiv}, pages arXiv--2004.

\bibitem[{Turing(2009)}]{turing1950computing}
Alan~M Turing. 2009.
\newblock Computing machinery and intelligence.
\newblock In \emph{Parsing the turing test}, pages 23--65. Springer.

\bibitem[{de~Vries et~al.(2020)de~Vries, Bahdanau, and
  Manning}]{de_vries_towards_2020}
Harm de~Vries, Dzmitry Bahdanau, and Christopher Manning. 2020.
\newblock Towards ecologically valid research on language user interfaces.
\newblock \emph{arXiv e-prints}, pages arXiv--2007.

\bibitem[{Wen et~al.(2016)Wen, Vandyke, Mrksic, Gasic, Rojas-Barahona, Su,
  Ultes, and Young}]{wen2016network}
Tsung-Hsien Wen, David Vandyke, Nikola Mrksic, Milica Gasic, Lina~M
  Rojas-Barahona, Pei-Hao Su, Stefan Ultes, and Steve Young. 2016.
\newblock A network-based end-to-end trainable task-oriented dialogue system.
\newblock \emph{arXiv preprint arXiv:1604.04562}.

\bibitem[{Williams et~al.(2017)Williams, Asadi, and
  Zweig}]{williamsHybridCodeNetworks2017}
Jason~D Williams, Kavosh Asadi, and Geoffrey Zweig. 2017.
\newblock Hybrid code networks: practical and efficient end-to-end dialog
  control with supervised and reinforcement learning.
\newblock \emph{arXiv preprint arXiv:1702.03274}.

\bibitem[{Zhang et~al.(2018)Zhang, Dinan, Urbanek, Szlam, Kiela, and
  Weston}]{zhangPersonalizingDialogueAgents2018}
Saizheng Zhang, Emily Dinan, Jack Urbanek, Arthur Szlam, Douwe Kiela, and Jason
  Weston. 2018.
\newblock Personalizing dialogue agents: I have a dog, do you have pets too?
\newblock \emph{arXiv preprint arXiv:1801.07243}.

\bibitem[{Zhang et~al.(2019)Zhang, Sun, Galley, Chen, Brockett, Gao, Gao, Liu,
  and Dolan}]{zhang2019dialogpt}
Yizhe Zhang, Siqi Sun, Michel Galley, Yen-Chun Chen, Chris Brockett, Xiang Gao,
  Jianfeng Gao, Jingjing Liu, and Bill Dolan. 2019.
\newblock Dialogpt: Large-scale generative pre-training for conversational
  response generation.
\newblock \emph{arXiv preprint arXiv:1911.00536}.

\bibitem[{Zhao et~al.(2017)Zhao, Zhao, and Eskenazi}]{zhao2017learning}
Tiancheng Zhao, Ran Zhao, and Maxine Eskenazi. 2017.
\newblock Learning discourse-level diversity for neural dialog models using
  conditional variational autoencoders.
\newblock \emph{arXiv preprint arXiv:1703.10960}.

\end{thebibliography}
\bibliographystyle{acl_natbib}

\clearpage
\newpage
\appendix

\section{Tasks}
\label{apx:tasks}

Table~\ref{tab:tasks} shows the 24 tasks and their descriptions.
\begin{table*}[ht]
\begin{tabular}{ll}
\textbf{Task name}            & \textbf{Description}                                                 \\
book\_apartment\_viewing      & Schedule an apartment viewing, given the name of the rental company  \\
apartment\_search             & Find an apartment to rent                                            \\
bank\_balance                 & Check the balance of a bank account (A).                             \\
bank\_fraud\_report           & Report suspicious behavior on your bank account (A).                \\
book\_doctor\_appointment     & Make an appointment with a doctor                                    \\
followup\_doctor\_appointment & Check instructions given by doctor upon last visit                   \\
hotel\_reserve                & Reserve a room in a hotel, given its name                            \\
hotel\_search                 & Find a hotel                                                         \\
hotel\_service\_request       & Ask for a room service in a hotel                                    \\
schedule\_meeting             & Schedule a meeting                                                   \\
party\_plan                   & Plan a party at a given venue                                        \\
party\_rsvp                   & RSVP to a party of a given host at a given venue                     \\
plane\_search                 & Find a flight between two cities                                     \\
plane\_reserve                & Book a flight, given its id                                          \\
restaurant\_search            & Find a restaurant                                                    \\
restaurant\_reserve           & Reserve a table at a restaurant                                      \\
book\_ride                    & Call a Taxi/Uber/Lyft ride to any destination                        \\
ride\_change                  & Change details of a Taxi/Uber/Lyft ride that had been called earlier \\
ride\_status                  & Check the status of a ride you called earlier                        \\
spaceship\_life\_support      & Recover the spaceship's life support                                 \\
spaceship\_access\_codes      & Get a repair robot to open a door for you                            \\
trip\_directions              & Get walking/driving/transit directions between two locations (B).    \\
trivia                        & Play a game of trivia (C).                                           \\
weather                       & Check the weather (forecast) in various cities               
\end{tabular}
\caption{Descriptions of the 24 tasks. Particular challenges are introduced by (A) different options for what information the user can provide (security questions in case of forgotten PIN number), (B) repeated information retrieval from a single knowledge base item, and (C) a loop in the schema with a summary of events in that loop at the end.}
\label{tab:tasks}
\end{table*}

\begin{figure*}[!htb]
    \centering
    \includegraphics{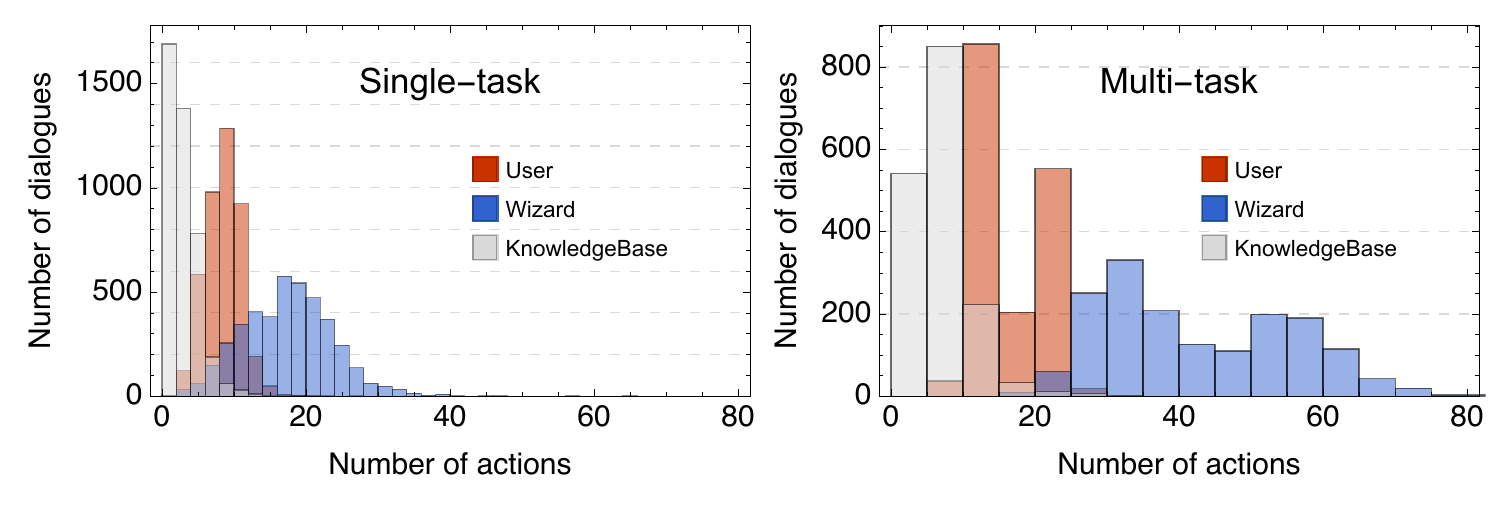}
    \caption{Distribution of action counts for single task (left panel) and multi-task (right panel) dialogs. }
    \label{fig:action_counts}
\end{figure*}

Figure~\ref{fig:action_counts} shows the distribution of action counts for single- and multi-task dialogs, respectively.
For multi-task setups, action counts double for the user and the wizard and exhibit a substantial increase in knowledge base queries, highlighting how difficulty and complexity increase when moving from a single-task to a multi-task setup, which typically cover 2-5 tasks.

\begin{figure*}[ht]
    \centering
    \includegraphics[width=1.00\textwidth]{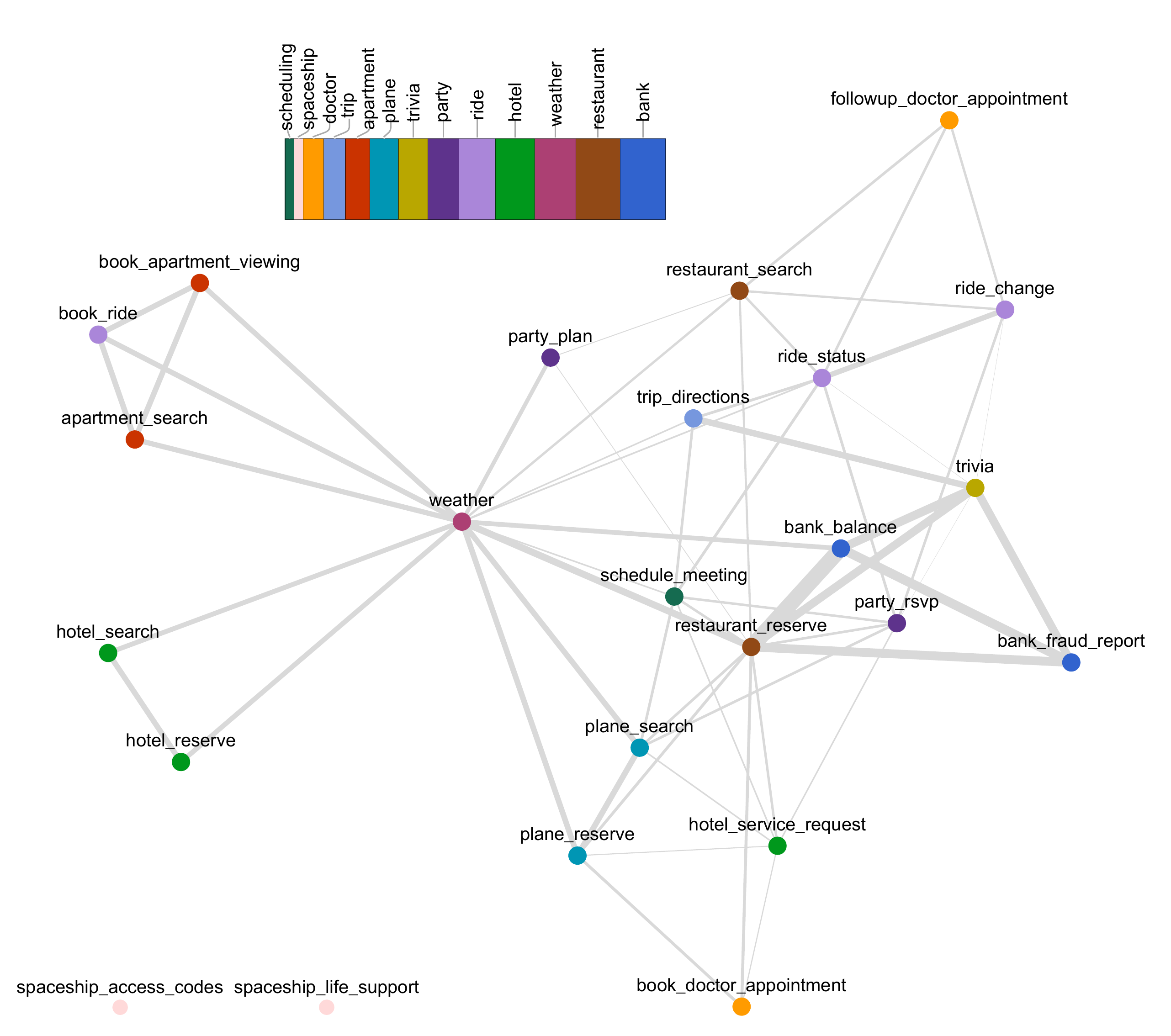}
    \caption{%
      Co-occurrence of tasks.
      Each task is shown as a vertex in this graph.
      The more dialogs two tasks jointly appear in (based on the knowledge base queries in that dialog) the thicker the edge connecting these tasks.
      Some tasks do not appear together in any dialog and are therefore not connected.
      Vertex colors indicate the task domain.
      The relative frequency of the 13 domains is illustrated in the top left inset.
      The largest clique in this graph is \{restaurant\_reserve, schedule\_meeting, plane\_search, hotel\_service\_request, party\_rsvp\}.
    }
    \label{fig:task_relations}
\end{figure*}

\begin{figure*}[tb]
    \centering
    \includegraphics[width=1.00\textwidth]{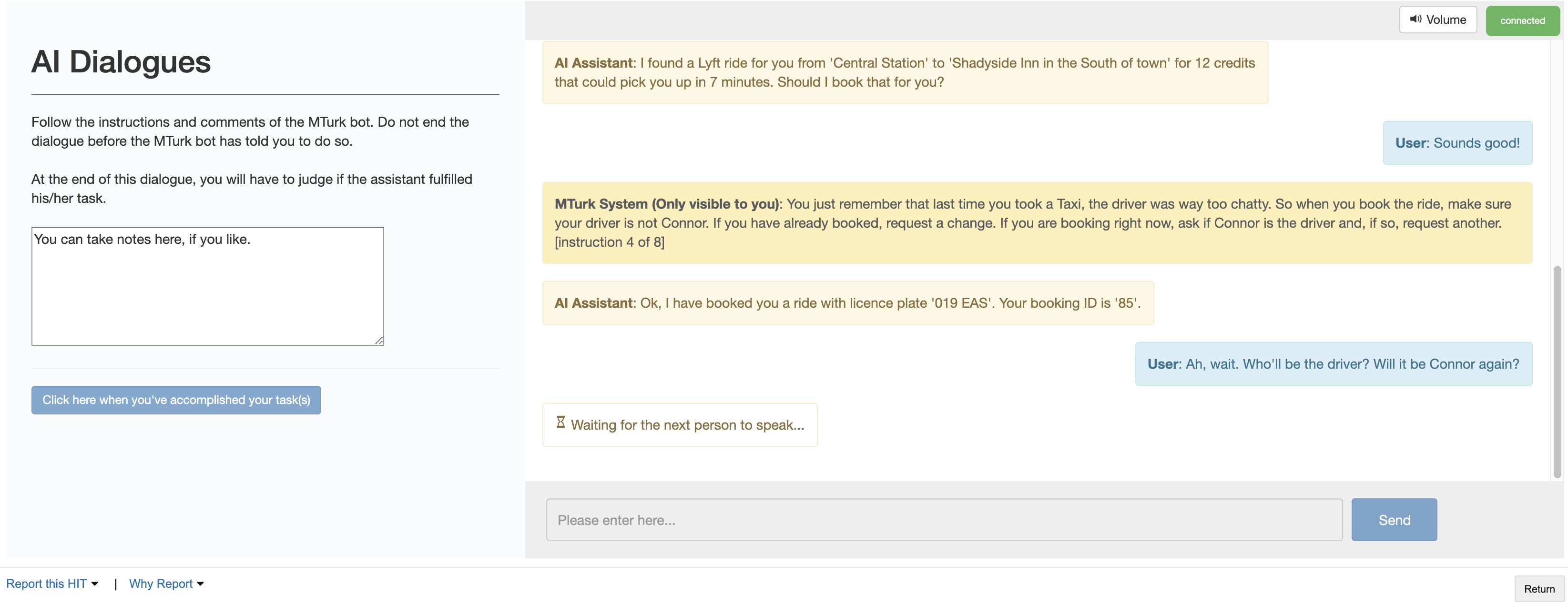}
    \caption{%
      User's graphical user interface. %
      The left panel shows general instructions and a note-taking area. %
      The right panel shows the dialog, where during-dialog instructions are shown as dark-yellow boxes, AI Assistant (wizard) messages as light-yellow boxes, and the user's messages as blue boxes.
    }
    \label{fig:ui_user}
\end{figure*}

Multi-task scenarios connect tasks as is shown in Figure~\ref{fig:task_relations}.
For the user, task instructions are sometimes given during the dialog, as shown in Figure~\ref{fig:ui_user}.

Table~\ref{tab:correctly_ordered_fractions} shows the fraction of single-task dialogs per task in which the wizards follow the prescribed order of questions at the beginning of the dialog.
\begin{table*}[ht]
\begin{tabular}{|llll|}
\hline
Task & Happy & All & Number of Checked Actions \\
\hline
apartment\_schedule      & 0.95 & 0.94 & 5 \\
apartment\_search        & 0.90 & 0.90 & 7 \\
bank\_balance            & 0.97 & 0.97 & 3 \\
bank\_fraud\_report      & 0.97 & 0.96 & 3 \\
doctor\_followup         & 0.98 & 0.98 & 4 \\
doctor\_schedule         & 0.90 & 0.90 & 6 \\
hotel\_book              & 0.95 & 0.91 & 6 \\
hotel\_search            & 1.00 & 1.00 & 5 \\
hotel\_service\_request  & 0.66 & 0.66 & 6 \\
meeting\_schedule        & 0.94 & 0.94 & 6 \\
party\_plan              & 0.77 & 0.77 & 6 \\
party\_rsvp              & 0.67 & 0.67 & 7 \\
plane\_book              & 0.97 & 0.94 & 3 \\
plane\_search            & 1.00 & 1.00 & 4 \\
restaurant\_book         & 0.84 & 0.84 & 5 \\
restaurant\_search       & 0.99 & 0.99 & 7 \\
ride\_book               & 0.98 & 0.95 & 4 \\
ride\_change             & 0.92 & 0.92 & 4 \\
ride\_status             & 0.90 & 0.90 & 3 \\
spaceship\_access\_codes & 0.88 & 0.88 & 5 \\
spaceship\_life\_support & 0.99 & 0.95 & 4 \\
trip\_directions         & 0.79 & 0.76 & 5 \\
trivia                   & 1.00 & 1.00 & 2 \\
weather                  & 1.00 & 1.00 & 2 \\
\hline
mean                     & 0.91 & 0.91 & \\
\hline
\end{tabular}
\caption{%
Fraction of single-task dialogs per task in which the wizards follow the prescribed order of questions at the beginning of the dialog. %
Here, skipping questions or repeating them is allowed, as the user can provide multiple pieces of information at once or say something irrelevant. %
The fractions shown in the second column concern happy-path dialogs only, whereas the third column includes all single-task dialogs. %
The last column shows the number of actions that are checked for correct order. %
For example, in the plane\_book task, the three actions ``ask\_name'', ``plane\_ask\_flight\_id'', and ``query plane\_book'', must occur in that order. %
}
\label{tab:correctly_ordered_fractions}
\end{table*}

\section{Worker payments}
\label{apx:payment}

Here we summarize the payments and bonuses that workers receive for their work.

In Stage~I, workers that use fewer than 9 hints to solve the questionnaire are paid \$2 for participating in the tutorial, otherwise their assignment is rejected. 
In addition, successful workers are paid a bonus of \$0.50 or \$0.25 if they answer without using hints or with up to three hints, respectively.

In a second evaluation step, we consider workers for which both initially considered dialogs where acceptable, and who have submitted at least 10 assignments for each role (user and wizard).
We manually read samples of their submissions and decide who are members of the top performing 10\%, based on how well they followed their instructions and how creative they've been as a user.
Those workers get paid a bonus amount of \$0.40 per assignment. 

The payment in Stage~III is identical to that of Stage~I, but the video is shorter and, therefore, the hourly salary is higher.

In Stage~IV we pay \$1.90 per HIT plus a bonus of \$0.90 cents for particularly long instruction sets (they take about 20 minutes to complete). 
Once again, the top 10\% of workers get paid a bonus of \$0.60 per assignment.

\section{Data format}
\label{apx:format}

Each dialog is stored as a JSON file with the following structure.
\small
\begin{verbatim}
{
  "FORMAT-VERSION": 6,
  "dialogID": integer number of 
                the dialog,
  "BatchID": string identifier for 
             the collection batch,
  "CompletionLevel": string 
             indicating possible 
             disconnects,
  "Scenario": {
    "Domains": list of domains 
               that this dialog 
               can touch on,
    "UserTask": initial instructions 
                given to the user,
    "WizardTask": instructions given 
                to the wizard,
    "WizardCapabilities": list of 
                capabilities of the 
                wizard (see below),
    "Happy": boolean indicating if 
                this scenario was 
                happy,
    "MultiTask": boolean indicating 
                if the wizard had 
                more than one task 
                capability,
  },
  "Events": list of events that 
            constitute the dialog,
  "AnonymizedWizardWorkerID": string 
            identifying the wizard 
            worker,
  "AnonymizedUserWorkerID": string 
            identifying the user 
            worker,
  "UserQuestionnaire": list of 
            questions and answers 
            for the user at the end 
            of the dialog,
  "WizardQuestionnaire": list of 
            questions and answers 
            for the wizard at the 
            end of the dialog,
}
\end{verbatim}
\normalsize
At the heart of the dialog file is the list of events.
Events can be user utterances or the custom response of the wizard, but also one of the other things that we describe in Table~\ref{tab:events}.
Apart from time stamp data, all events have an \verb+Agent+ and an \verb+Action+ field.
Further information stored with the event depends on these latter two fields, as can be seen in Table~\ref{tab:events}.
\begin{table*}[ht]  
    \begin{tabular}{lll}
        \textbf{Agent} & \textbf{Action}      & \textbf{Additional information}                        \\
        User           & utter                & Text                                                   \\
        User           & complete             &                                                        \\
        Wizard         & request\_suggestions & Text, PrimaryItem, SecondaryItem                       \\
        Wizard         & pick\_suggestion     & Text, Itent, IntentOptions, PrimaryItem, SecondaryItem \\
        Wizard         & utter                & Text, PrimaryItem, SecondaryItem                       \\
        Wizard         & query                & Constraints, API, PrimaryItem, SecondaryItem           \\
        Wizard         & select\_topic        & Topic, PrimaryItem, SecondaryItem                      \\
        Wizard         & select\_primary      & PrimaryItem, SecondaryItem                             \\
        Wizard         & select\_secondary    & PrimaryItem, SecondaryItem                             \\
        KnowledgeBase  & return               & Item, TotalItems, Topic                                \\
        UserGuide      & instruct             & Text                                                  
    \end{tabular}
    \cprotect\caption{%
      Information stored in dialog events. %
      Each event contains an Agent and an Action field. %
      Additional fields depend on the former two. %
      The Text field contains a plain text utterance of the agent. %
      The PrimaryItem and SecondaryItem fields contain JSON objects that describe the selected knowledge base item, and the knowledge base item that is selected for comparison. %
      Both of these fields may be \verb+null+. %
      The Intent field is a label associated with the utterance that the wizard has selected. %
      If the intent is \verb+custom+, then the wizard used a free-form reply. %
      The IntentOptions field is a list of suggested responses (given as intent labels) that the wizard could choose from. %
      The Constraints field is a JSON object describing the constraints on the knowledge base query. %
      Here, comparators may be expressed as python expressions, e.g. \verb+api.is_greater_than(4)+. %
      The API field is the name of the task that the event is related to.
      Finally, the Item field is a JSON object describing a knowledge base item and the TotalItems field is the number of knowledge base items that would also satisfy the specified constraints, or -1 if not applicable. %
    }
    \label{tab:events}
\end{table*}

The task schemas are also stored as JSON files.
For example, the schema of the \verb+book_doctor_appointment+ task is:
\small
\begin{verbatim}
{
  "task": "book_doctor_appointment",
  "replies": {
    "hello": "Hello, how can I help?",
    "ask_name": "Could I have your name, 
      please?",
    "doctor_ask_doctor_name": "What 
      doctor would you like to see?",
    "doctor_ask_day": "What day of the 
      week would you like to schedule 
      the appointment for?",
    "doctor_ask_start_time": "At what 
      time can you be at the clinic?",
    "doctor_ask_symptoms": "Could you 
      describe your symptoms, please?",
    "doctor_inform_booking_unavailable": 
      "Unfortunately {doctor_name:s} 
      has no appointment available at 
      {time:s}.",
    "doctor_inform_booking_available": 
      "Alright, {doctor_name:s} is 
      available at {time:s}. Can I book 
      the appointment for you?",
    "doctor_inform_booking_successful": 
      "Great, your appointment with 
      {doctor_name:s} is booked for you!",
    "doctor_bye": "Thank you and goodbye.",
    "doctor_inform_nothing_found": 
      "Unfortunately there is currently 
      no doctor available.",
    "anything_else": "Is there anything 
      else that I can do for you?",
    "out_of_scope": "I am sorry, I don't 
      quite understand what you mean. I am 
      only able to help you book an 
      appointment with your doctor.",
    "unavailable": "Unavailable",
    "available": "Available",
    "no": "No",
    "yes": "Yes",
    "query_check": "Query Check",
    "query_book": "Query Book"
  },
  "graph": {
    "hello": "ask_name",
    "ask_name": "doctor_ask_doctor_name",
    "doctor_ask_doctor_name": 
      "doctor_ask_day",
    "doctor_ask_day": 
      "doctor_ask_start_time",
    "doctor_ask_start_time": 
      "doctor_ask_symptoms",
    "doctor_ask_symptoms": "query_check",
    "available": 
      "doctor_inform_booking_available",
    "unavailable": 
      "doctor_inform_booking_unavailable",
    "yes": "query_book",
    "no": "doctor_ask_doctor_name",
    "query_book": 
      "doctor_inform_booking_successful",
    "doctor_inform_booking_successful": 
      "anything_else"
  }
}
\end{verbatim}
\normalsize

\section{Multiple-choice tests}
\label{apx:multiple_choice}

To prepare workers for the Stage~II and Stage~IV tasks, we ask them to watch video tutorials and answer the following questions during Stages I and III.

\subsection{Stage~I - Single Task}
\label{apx:multiple_choice:single}

The Stage~I video tutorial is available at \anonymous{\url{https://youtu.be/L7QpscLPTFM}}. 
After watching the video, workers have to answer the following questionnaire with fewer than 9 hints.
\begin{enumerate}
    \item As the user, when can you end the task? (Correct answer: (c))
    \begin{enumerate}
        \item I cannot. Only the assistant can do this.
        \item As soon as the 'Click here when you've accomplished your task(s)' button is enabled.
        \item When I have followed all the instructions, including those on the left panel and those that are given by the MTurk System Bot.
    \end{enumerate}
    \item As the user, if you forget to follow an instruction from the MTurk System bot, what should you do? (Correct answer: (b))
    \begin{enumerate}
        \item Forget about the instruction since it is outdated.
        \item Follow the instruction once I realise that I've overlooked it.
        \item Tell the AI Assistant that I've missed an instruction.
    \end{enumerate}
    \item As the user, what should you do if you need some information that was not given in the task instructions? (Correct answer: (a))
    \begin{enumerate}
        \item I can just make something up.
        \item Give up.
        \item I must avoid answering any questions about the missing information.
    \end{enumerate}
    \item As the assistant, what should you do if the user begins a dialog, but doesn't say what he/she wants? (Correct answer: (b))
    \begin{enumerate}
        \item I tell the user what task I can help her with
        \item I should just greet the user
        \item I should say that I cannot understand what he/she's saying
    \end{enumerate}
    \item For the assistant, what of these things is the MOST important? (Correct answer: (b))
    \begin{enumerate}
        \item Being helpful to the user
        \item Following the flow chart of the current task whenever possible
        \item Making the conversation as short as possible
        \item Making the conversation as long as possible
    \end{enumerate}
    \item What does the request-optional box mean in the flow chart? (Correct answer: (a))
    \begin{enumerate}
        \item It represents information that I can use, but that is not absolutely required.
        \item I can ignore this box if it doesn't make sense here
    \end{enumerate}
    \item As an assistant, when should you use one of the suggested responses? (Correct answer: (a))
    \begin{enumerate}
        \item Whenever possible - I only use custom responses if the situation is not accounted for in the flow chart
        \item Only if they fit exactly what I want to say
    \end{enumerate}
    \item What does the ``request type'' form field mean? (Correct answer: (a))
    \begin{enumerate}
        \item It is there to distinguish between checking if a booking is available or actually performing the booking
        \item It's about the type of person I am dealing with
        \item The request type describes the topic that the user is referring to.
    \end{enumerate}
    \item How does the ideal user behave? (Correct answer: (b))
    \begin{enumerate}
        \item As a user, I should behave in a structured and consistent manner. Always friendly and concise.
        \item As a user, I should make the dialog more interesting and complicated (but follow instructions).
        \item Just follow the instructions.
    \end{enumerate}
    \item How does the ideal AI Assistant behave? (Correct answer: (c))
    \begin{enumerate}
        \item Just follow the instructions.
        \item As an AI Assistant, I should make the dialog more complex and be very creative - but follow the instructions.
        \item As an AI Assistant, I should behave in a structured and consistent manner. Always friendly and concise.
    \end{enumerate}
\end{enumerate}

\subsection{Stage~III - Multi Task}
\label{apx:multiple_choice:multi}

The Stage~III video tutorial is available at \anonymous{\url{https://youtu.be/dd0s2Sqox6g}}. 
After watching the video, workers have to answer the following questionnaire with fewer than 9 hints.
\begin{enumerate}
    \item Which one of these three things has changed, compared to the single-task dialogs? (Correct answer: (a))
    \begin{enumerate}
        \item The number of flow-charts that the AI Assistant has to follow.
        \item The fact that one plays either a user or an AI Assistant.
        \item The design of the AI Assistant's flow-charts.
    \end{enumerate}
    \item Why is it important to use names (for places, airlines, etc.) that are given in the left-panel instructions (if any)? (Correct answer: (b))
    \begin{enumerate}
        \item Those names are better.
        \item The AI Assistant may only be able to choose from a limited set of names, so if I come up with something random, the assistant cannot help me.
        \item This is a complicated issue that has to do with machine learning.
    \end{enumerate}
    \item As the user, what should you do if you need some information that was NOT given in the task instructions? (Correct answer: (a))
    \begin{enumerate}
        \item I can just make something up.
        \item Give up.
        \item I must avoid answering any questions about the missing information.
    \end{enumerate}
    \item As the assistant, what should you do if the user begins a dialog, but doesn't say what he/she wants? (Correct answer: (c))
    \begin{enumerate}
        \item I tell the user what task I can help her with
        \item I should say that I cannot understand what he/she's saying
        \item I should just greet the user
    \end{enumerate}
    \item For the assistant, what of these things is the MOST important? (Correct answer: (b))
    \begin{enumerate}
        \item Being helpful to the user
        \item Following the flow chart of the current task whenever possible
        \item Making the conversation as short as possible
        \item Making the conversation as long as possible
    \end{enumerate}
    \item If two or more topics are given, and you (as a user) are otherwise free to form the dialog, what would NOT be ok? (Correct answer: (c))
    \begin{enumerate}
        \item Jumping back and forth between topics.
        \item Occasionally saying something that's not really relevant to the conversation (smalltalk).
        \item Covering only one of the given topics.
    \end{enumerate}
    \item As an assistant, when should you use one of the suggested responses? (Correct answer: (a))
    \begin{enumerate}
        \item Whenever possible - I only use custom responses if the situation is not accounted for in the flow chart
        \item Only if they fit exactly what I want to say
    \end{enumerate}
    \item As an assistant, what should you do when the user changes tasks? (Correct answer: (b))
    \begin{enumerate}
        \item I switch to the corresponding task-tab and follow this task's flow chart from the beginning
        \item I switch to the corresponding task-tab and follow this task's flow chart from the point that makes most sense, given what information I already have.
        \item I use a free-form response.
    \end{enumerate}
    \item How does the ideal user behave? (Correct answer: (b))
    \begin{enumerate}
        \item As a user, I should behave in a structured and consistent manner. Always friendly and concise.
        \item As a user, I should make the dialog more interesting and complicated (but follow instructions).
        \item Just follow the instructions.
    \end{enumerate}
    \item How does the ideal AI Assistant behave? (Correct answer: (c))
    \begin{enumerate}
        \item Just follow the instructions.
        \item As an AI Assistant, I should make the dialog more complex and be very creative - but follow the instructions.
        \item As an AI Assistant, I should behave in a structured and consistent manner. Always friendly and concise.
    \end{enumerate}
\end{enumerate}

\onecolumn
\section{Example Dialogs}
\label{apx:example_dialogs}

What follows are example dialogs from the three classes: happy, unhappy, and multi-task.
Here, we only display the plain text utterances exchanged by the user and the wizard and the knowledge base items received by the wizard.
We suppress the knowledge base queries and other annotation.

\newcommand{\usr}[1]{%
\noindent USR: {#1} \newline %
}
\newcommand{\wiz}[1]{%
\noindent WIZ: {#1} \newline %
}
\newcommand{\kbq}[1]{%
\noindent {\color{gray}KB\phantom{X}: {#1}} \newline %
}

\subsection{Single-Task, Happy}

\usr{I would like to make a service request for 6 am }
\wiz{Could I get your name, please?}
\usr{Mark}
\wiz{At what hotel are you currently staying?}
\usr{Old Town Inn room 359}
\wiz{Right, please let us know your request now.}
\usr{I want to order medium rare steak and a glass of red wine}
\kbq{RequestStatus: Request Confirmed, RoomNumber: 359, Time: 6 am, api\_name: hotel\_service\_request}
\wiz{Your request has been submitted successfully.
A member of the service team will knock on the door of room 359 at 6 am!}
\usr{okay thanks}
\wiz{Is there anything else that I can do for you?}

\subsection{Single-Task, Unhappy}

\usr{Hi i need a room at the Hilton please}
\wiz{What hotel would you like to stay at?}
\usr{Ah actually my friend just told me Old Town Inn is better, i whant that one}
\wiz{May I have your name, please?}
\usr{Yes its Mark, }
\wiz{When are you arriving?}
\usr{Ah he was saying the Old Town is better in naother city, i just saw the Shadyside inn online and think its better, May7th to 23rd please}
\kbq{Message: Unavailable, HotelName: Shadyside Inn, api\_name: hotel\_reserve}
\wiz{I'm very sorry, but there is no room available at the Shadyside Inn for your requested dates.}
\usr{Oh no, this blog never let me down, its so great usually, can you check again?}
\kbq{Message: Available, HotelName: Shadyside Inn, api\_name: hotel\_reserve}
\wiz{Alright, the Shadyside Inn ticks all of your boxes, can I book this room for you?}
\usr{Ah, cant beleve this i just got invited to a wedding in LONDON, lol please change the dates to 15th-26th the following month}
\kbq{Message: Unavailable, HotelName: Shadyside Inn, api\_name: hotel\_reserve}
\wiz{I'm very sorry, but there is no room available at the Shadyside Inn for your requested dates.}
\usr{Oh, bad news all the way, ok thats it i guess}
\wiz{Would you like to try searching for another Hotel?}
\usr{No thats it, thanks bye!}
\wiz{Thank you and goodbye.}

\subsection{Multi-Task}

\vspace{1em}
\small
\usr{Hey, I need to change a ride I have already booked!}
\wiz{Could you give me your name, please?}
\usr{Ben}
\wiz{Can I get your booking ID, please?}
\usr{It is ride id 551.}
\wiz{Sure, what can I change for you?}
\usr{I need to change the arrival time if possible. I won't be ready for at least 20 more minutes.}
\kbq{ChangeStatus: We are unable to change your trip., api\_name: ride\_change}
\wiz{Unfortunately I wasn't able to update your booking, sorry.}
\usr{Oh no! Well they might have to wait. Can you give me the status of the ride now please?}
\kbq{RideStatus: Your ride is on its way., RideWait: 3 minutes away, api\_name: ride\_status}
\wiz{Your ride is 3 minutes away}
\usr{Yikes! I am sick on my stomach and I think it is because of these meds Dr. Morgan gave me. Can you check if I was supposed to take this with food, or only water, or what?}
\kbq{Message: You must take your medicine 2 times a day before meals., api\_name: followup\_doctor\_appointment}
\wiz{Your instructions are: You must take your medicine 2 times a day before meals..}
\usr{Oh, I didn't eat with it. I bet that is why I am sick now.}
\wiz{Is there anything else that I can do for you?}
\usr{I will need to find a restaurant to go to dinner, so I will have something to eat before my next scheduled dose.}
\wiz{What restaurant would you like to go to?}
\usr{I don't have a real preference. I would like something in the North or South of town though.}
\wiz{What cuisine are you looking for?}
\usr{I would like pizza or Italian. But I am open. Just anything but Chinese.}
\kbq{id: 366, Name: The Porch, Cost: Cheap, TakesReservations: False, DoesDelivery: False, AverageRating: 2, Food: Pizza, AverageWaitMinutes: 31, OpenTimeHour: 7 am, CloseTimeHour: 9 pm, MaxPartySize: 9, Location: West, api\_name: restaurant\_search}
\wiz{Any minimum rating that the restaurant should have?}
\usr{No, sometimes ratings are set up. You know people get their friends to do their Yelp and Google reviews.}
\kbq{id: 219, Name: Legume, Cost: Moderate, TakesReservations: True, DoesDelivery: False, AverageRating: 5, Food: Steak, AverageWaitMinutes: 50, OpenTimeHour: 11 am, CloseTimeHour: 7 pm, MaxPartySize: 6, Location: North, api\_name: restaurant\_search}
\wiz{I found Legume restaurant in the North side. There is an average wait time of 50 minutes and it is moderately priced.}
\usr{That is a wait. My aunt is diabetic. Can you search for something else with less of a wait time?}
\kbq{id: 884, Name: The Porch, Cost: Moderate, TakesReservations: False, DoesDelivery: True, AverageRating: 5, Food: Italian, AverageWaitMinutes: 27, OpenTimeHour: 6 am, CloseTimeHour: 11 pm, MaxPartySize: 42, Location: North, api\_name: restaurant\_search}
\wiz{The Porch Italian restaurant is in the North side. There is an average wait of 27 minutes and it is moderately priced.}
\usr{Ok, I will tell her to bring her sugar tablets just in case. But that should be acceptable.}
\wiz{Is there anything else that I can do for you?}
\usr{You know what? I forgot that I saw Dr. Alexis last week too and have pills from her. Can you check those instructions for me? One might be with food and the other not. Who knows?}
\kbq{Message: You must take your medicine 2 times a day before meals., api\_name: followup\_doctor\_appointment}
\wiz{Your instructions are: You must take your medicine 2 times a day before meals..}
\usr{Ok, they have the same instructions, that's a relief. I have enough to think about.}
\wiz{Is there anything else that I can do for you?}
\usr{Is my ride still 3 minutes away? He hasn't arrived here yet?}
\kbq{RideStatus: Your driver is dropping off another passenger., RideWait: 9 minutes away, api\_name: ride\_status}
\wiz{Your driver is dropping off another passenger and is 9 minutes away.}
\usr{Oh, ok. That actually works out. I am still on the toilet. Smh. I will definitely eat after I take it next time. }
\wiz{Is there anything else that I can do for you?}
\usr{I just want to tell you that you have been most helpful to me today.}
\wiz{Thank you and goodbye.}
\usr{Thank you so much!}
\wiz{Goodbye. Enjoy your ride!}
\normalsize

\twocolumn
\section{Selected Comments from Turkers}
\label{apx:turker_comments}

We received comments such as those below throughout the four stages of data collection.
We also received emails with detailed constructive (though mostly positive) feedback that spans up to an entire page, which we do not share here.
The primary negative feedback concerned partner disconnects during the dialogs and we intend to mitigate this in future projects.
\vspace{1em}

\newcommand{\q}[2]{%
\noindent{\textit{``#1''}}%
\begin{flushright}%
{#2}%
\end{flushright}%
\vspace{1em}%
}

\q{Thank you for doing such a nice job.}{P. M.}
\q{Will there be more hits posted? I love your hits!}{S. K.}
\q{I just wanted to take this opportunity to send out a warm thank you for the work you recently have posted. I hope the data you are getting will meet or excel your expectations and if there are any concerns or issues with dialog or extended conversations that you need, please don't hesitate to contact me. Please stay safe during these times and I look forward to many more batches of work and working for you in the future. Thanks again, have a great day!}{B. C.}
\q{Thanks for everything! I'm loving participating.}{T. R.}
\q{I thoroughly enjoyed taking part in this study.}{I. B. P.}
\q{I loved doing your dialogs, they were so much fun and engaging.}{N. W.}
\q{Will there be any work today? Looking forward to the next round :)}{L. H.}
\q{I liked how clearly you explained the directions and that you made clear what your expectations were. The training videos were also very helpful, they made it easier for me to learn quickly and to be able to perform the HITs up to your standards. I also think the tests were helpful too because they really brought to my attention what was important in the video. For example, one of the test questions the early stage asked about whether the AI's role was to bring the conversation back to the script or to make sure the customer was satisfied. The answer really let me know what my role would be, which I liked. %
\newline %
My only dislike was that during stage IV I occasionally had partners who were obviously working on more than one task at once, which led to long delays for simple responses. This became very frustrating when HITs would drag on for a long time due to inactivity from partners. I prefer to focus my full attention on each HIT, so I did become a bit impatient at times. However, I understand that there is only so much you can control so I do understand why this happens.}{C. S.} 
\q{I would definitely like to work with you again in the future, I really enjoyed doing these tasks and can't wait for future projects!}{L. J.} 
\q{I enjoyed working on your project and am sorry to see it come to a close. I'd love to help out with future projects you might have. Please keep me in mind if you need a beta tester. %
\newline %
I'd also be interested in reading about your research if/when you publish your findings. 
Thanks again for the HITs; it was a pleasure working for you guys.}{A. F.} 

\end{document}